\documentclass{article} % For LaTeX2e
\usepackage{iclr2027_conference,times}

\usepackage[utf8]{inputenc}
\usepackage[T1]{fontenc}
\usepackage{hyperref}
\usepackage{url}
\usepackage{booktabs}
\usepackage{amsfonts}
\usepackage{nicefrac}
\usepackage[protrusion=true,expansion=true,activate={true,nocompatibility},final,tracking=true,kerning=true,spacing=true,disable=footnote]{microtype}
\usepackage{xcolor}

\PassOptionsToPackage{numbers}{natbib}
\usepackage{graphicx,subcaption}
\usepackage{enumitem}
\usepackage{algorithm}
\usepackage{algorithmic}
\usepackage{booktabs}
\usepackage{ragged2e}
\usepackage{wrapfig}
\usepackage{arydshln}
\usepackage{setspace}
\usepackage{adjustbox}
\usepackage{framed}
\usepackage{wrapfig}

\usepackage{hyperref}
\usepackage{url}
\usepackage[table]{xcolor}
\usepackage{graphicx} 
\usepackage{tabularx} 
\usepackage{amssymb} 
\usepackage{booktabs} 
\usepackage{wrapfig}
\usepackage{enumitem}
\usepackage[normalem]{ulem}
\usepackage{subcaption}
\usepackage{makecell}
\usepackage{multirow}
\usepackage{capt-of}% or \usepackage{caption}
\usepackage{booktabs}
\usepackage{varwidth}
\usepackage{longtable}

\usepackage{pgfplots}
\usepackage{pgfplotstable}
\pgfplotsset{compat=1.18}
\usepgfplotslibrary{groupplots}

\usepackage{amssymb}
\usepackage{mathtools}
\usepackage{amsthm}
\usepackage{multirow}
\usepackage{xparse}

\usepackage{pgfplots}
\usepgfplotslibrary{polar,groupplots}
\pgfplotsset{compat=1.18}

\usepackage{pgfplots}
\usepackage{subcaption}
\usepackage{xcolor}

\usepackage{empheq}
\usepackage[most]{tcolorbox}
\usepackage{calc}

\newtcbox{\eqbox}{
  colback=blue!5, colframe=blue!5, boxrule=0pt, arc=3pt,
  nobeforeafter, math upper, tcbox raise base,
  enhanced, boxsep=2pt, left=0pt, right=0pt, top=0pt, bottom=0pt
}

\pgfplotsset{compat=1.18}

\usepackage{pgfplots}
\usepackage{subcaption}
\usepackage{xcolor}
\usepackage{pgfplotstable}
\pgfplotsset{compat=1.18}

\definecolor{m1}{RGB}{160,160,160}
\definecolor{m2}{RGB}{130,146,166}
\definecolor{m3}{RGB}{97,148,147}
\definecolor{m4}{RGB}{92,124,189}
\definecolor{m5}{RGB}{38,69,120}
\definecolor{m6}{RGB}{180,120,90}
\definecolor{m7}{RGB}{120,170,110}
\definecolor{m8}{RGB}{150,110,170}

\usepackage{xparse}
\newlength{\meanw}
\newlength{\stdw}

\newcommand{\code}{\url{https://github.com/MathEXLab/ATEMPO}} %TODO

\theoremstyle{plain}

\theoremstyle{remark}

\hypersetup{hidelinks}

\usepackage[most]{tcolorbox}
\usepackage{xcolor}

\definecolor{problembg}{RGB}{238,246,253}
\definecolor{problemrule}{RGB}{38,92,148}

\usepackage{amsthm}
\usepackage[most]{tcolorbox}
\usepackage{xcolor}

\definecolor{problemaccent}{RGB}{28,66,105}
\definecolor{problemtint}{RGB}{234,241,248}
\newtcolorbox{problemframe}{
  enhanced,
  breakable,
  left=11pt, right=9pt, top=7pt, bottom=7pt,
colback=problemtint, colframe=problemtint, boxrule=0pt, arc=3pt, left=2pt
}

\theoremstyle{definition}
\newtheorem*{problemthm}{Problem}

\newenvironment{problem}[1][]
  {\begin{problemframe}\begin{problemthm}[#1]}
  {\end{problemthm}\end{problemframe}}

\newtcolorbox{lossbox}{
colback=blue!5, colframe=blue!5, boxrule=0pt, arc=3pt, left=2pt, right=2pt, top=1pt, bottom=1pt, boxsep=0pt
}

\newtcolorbox{contributionbox}{
enhanced,
breakable,
left=11pt, right=9pt, top=7pt, bottom=7pt,
colback=green!5,colframe=green!5!,boxrule=0pt, arc=3pt, left=2pt
}

\definecolor{titlebg}{HTML}{E4E9F0}  % C05 pale-navy (current)
\definecolor{linkblue}{HTML}{116E8A}
\newcommand{\nuslogoheight}{0.80cm}
\newcommand{\mitlogoheight}{0.85cm}
\newcommand{\kuleuvenlogoheight}{1.33cm}
\newcommand{\logorowheight}{1.33cm}

\makeatletter
\newcommand{\fairauthors}[1]{\gdef\@fairauthors{#1}}
\newcommand{\fairaffiliations}[1]{\gdef\@fairaffiliations{#1}}
\newcommand{\fairabstract}[1]{\gdef\@fairabstract{#1}}
\newcommand{\faircorrespondence}[1]{\gdef\@faircorrespondence{#1}}
\gdef\@fairauthors{}\gdef\@fairaffiliations{}
\gdef\@fairabstract{}\gdef\@faircorrespondence{}

\newcommand{\synibmaketitle}{%
  \thispagestyle{empty}%
  \begingroup
  \tcbset{enhanced,frame hidden,
          left=0.7cm,right=0.7cm,top=0.32cm,bottom=0.28cm,
          arc=5pt,colback=titlebg,
          before skip=0pt,after skip=0.3cm,
          grow to left by=1.5pt,grow to right by=1.5pt}
  \begin{tcolorbox}
    \setlength{\parindent}{0pt}%
    \raggedright
    {\Large\bfseries \@title\par}
    \vskip 0.16cm
    {\small\bfseries \@fairauthors\par}
    \vskip 0.06cm
    {\small \@fairaffiliations\par}
    \vskip 0.16cm
    {\normalsize \@fairabstract\par}
    \vskip 0.16cm
    \noindent
    \begin{minipage}[c]{0.30\linewidth}
      \raggedright
      \ifx\@faircorrespondence\@empty\else
        {\small\textbf{Correspondence:} \@faircorrespondence}%
      \fi
    \end{minipage}%
    \hfill
    \begin{minipage}[c]{0.68\linewidth}
      \raggedleft
        \parbox[c][\logorowheight][c]{\widthof{\includegraphics[height=\nuslogoheight]{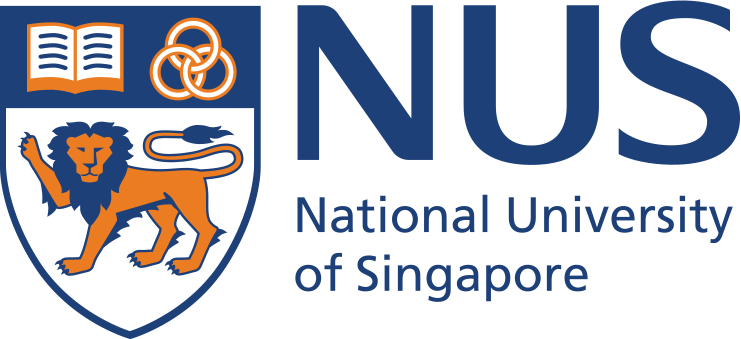}}}{%
          \centering\includegraphics[height=\nuslogoheight]{figures/logos/nus.png}%
        }%
        \hspace{0.15cm}%
        \parbox[c][\logorowheight][c]{\widthof{\includegraphics[height=\kuleuvenlogoheight]{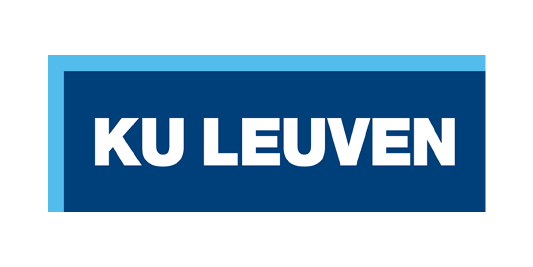}}}{%
          \centering\includegraphics[height=\kuleuvenlogoheight]{figures/logos/KU-Leuven-logo.png}%
        }%
        \hspace{0.1cm}%
      \mbox{%
        \parbox[c][\logorowheight][c]{\widthof{\includegraphics[height=\mitlogoheight]{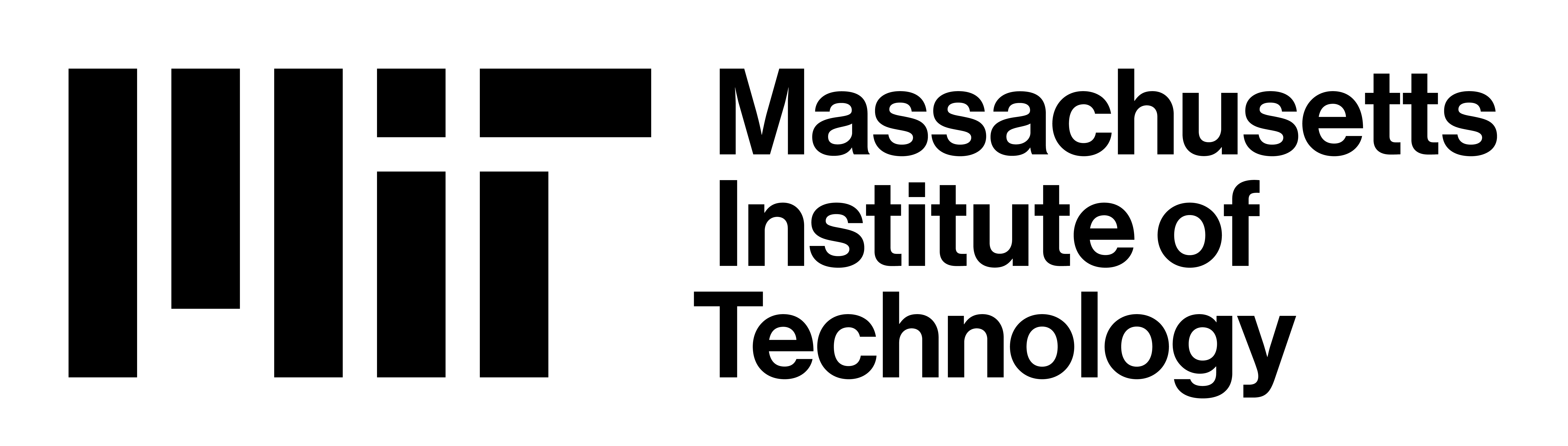}}}{%
          \centering\includegraphics[height=\mitlogoheight]{figures/logos/mit_lockup_std-three-line_rgb_black.png}%
        }%
      }%
    \end{minipage}%
  \end{tcolorbox}
  \endgroup
}
\makeatother
\title{\centering \LARGE ATEMPO: Amortized Transfer Entropy Model Per-lag Over-Time}
\author{}

\fairauthors{%
\centering 
\normalsize
Emma Andrews$^{1}$, 
Konstantinos Kontras$^{2,3}$,
Keane Ong$^{1,3}$, \\
Maarten De Vos$^{2}$,
Gianmarco Mengaldo$^{1}$
}
\fairaffiliations{%
\small
\centering 
  $^{1}$ National University of Singapore,\quad
  $^{2}$KU Leuven,\quad
  $^{3}$MIT,\quad}
\faircorrespondence{\href{mailto:emma_andrews@nus.edu.sg}{\texttt{emma\_andrews@nus.edu.sg}}}

\fairabstract{%
When one time series informs another, the delay at which that information arrives determines if models can utilize this relationship effectively. 
A source may be strongly predictive at one delay but uninformative at neighboring lags. Therefore, fixed input windows may obscure this structure by bundling informative and irrelevant history together. Transfer entropy (TE) can recover these delays by measuring how much a source's past tells us about a target beyond the target's own history. However, existing TE estimators require separate fits across source and conditioning depths, with further estimation on local windows to track delays over time. 
We introduce ATEMPO, a diffusion-based transfer entropy estimator amortized over lags and time. 
ATEMPO samples source and conditioning depths during training and conditions explicitly on temporal position, allowing a single score network to estimate the full TE surface without refitting. Differencing this surface recovers when information arrives.
Across synthetic datasets with ground-truth TE, ATEMPO achieves the lowest mean relative error among baselines, and on datasets with known lags, ATEMPO achieves \(0.98\pm0.04\) lag-detection accuracy. ATEMPO requires a single fit across source depths \(K=6\) and target depths \(L=3\), whereas existing estimators require 18 separate fits to evaluate the same \(K \times L\) grid. On three systems with time-varying lag, ATEMPO is the only estimator to maintain perfect lag-detection accuracy across all evaluated window lengths.
These recovered delays improve downstream forecasting: ATEMPO-selected inputs reduce MSE on WEATHER-5K by 7.25\%, while 120 selected inputs outperform the benchmark's 240-feature default by 4.6\%. Across three brain–computer interface tasks, using ATEMPO-selected lags achieves the highest or equal-best \(R^2\), improving mean \(R^2\) by 4.1\% relative to the strongest competing baseline.

}

\iclrfinalcopy % Uncomment for camera-ready version, but NOT for submission.

\begin{document}

\synibmaketitle

\section{Introduction}
\label{sec:intro}
Multivariate time-series models must decide which variables and past observations to expose to the model. In practice, this is often handled by applying a fixed-width window uniformly to every variable, selecting its length and offset using a validation sweep \citep{nie2023patchtst, liu2024itransformer, weather5k}. 
Yet a variable may become informative only at a particular delay, the gap between when it is observed and when its influence arrives. If that delay is thirty steps, a window long enough to reach it also includes the twenty-nine intervening lags, most of which carry nothing. Although an ideal model could learn to ignore irrelevant history, with finite data these additional inputs increase estimation difficulty and can obscure the value of the delayed signal.
This problem compounds in high-dimensional settings, where long common windows expose many potentially uninformative lags. Window selection therefore conflates two questions: \textit{whether} a variable is informative and \textit{when} its information arrives. Identifying the delay directly would allow a model to receive the informative observations without requiring all of the history in between. Transfer entropy (TE) \citep{schreiber2000} provides a natural way to identify such delays: it measures whether a source's past provides information about a target beyond that already contained in the target's own history.
%Window selection asks how much history to expose, delay discovery asks where within that history the information actually lies.
%The delay should therefore be measured in the data, before any model is trained. Take one candidate input and the series we want to predict. Given what that series' own past already tells us about its present, does the candidate, observed $k$ steps earlier, tell us any more? Correlation is symmetric and cannot answer this as it does not distinguish information already contained in the target's own history. 

The challenge is that delay identification requires more than one TE estimate, and repeated fitting is costly. Both the source depth \(k\) and conditioning depth \(l\) are unknown, and changing relationships must also be resolved over time \(t\). The target is therefore a surface \(TE(k,l,t)\), where existing estimators typically resolve individual points on this surface.
Tracking changing delays compounds the problem, as smaller local windows give better temporal resolution but fewer samples from which to estimate TE. Nearest-neighbor methods need exponentially many samples as the conditioning dimension grows \citep{gao2015}; variational bounds can become loose at large mutual information\citep{belghazi2018mutual, mcallester2020}; and cross-entropy estimators escape that bound but must discretize the continuous series \citep{mukherjee2020ccmi, shalev2024}. Diffusion-based estimators \citep{munoz2025} avoid these restrictions, however, they still require repeated fits as source and conditioning depths must be fixed before training.
\begin{figure}[t]
\begin{center}
\includegraphics[width=\textwidth]{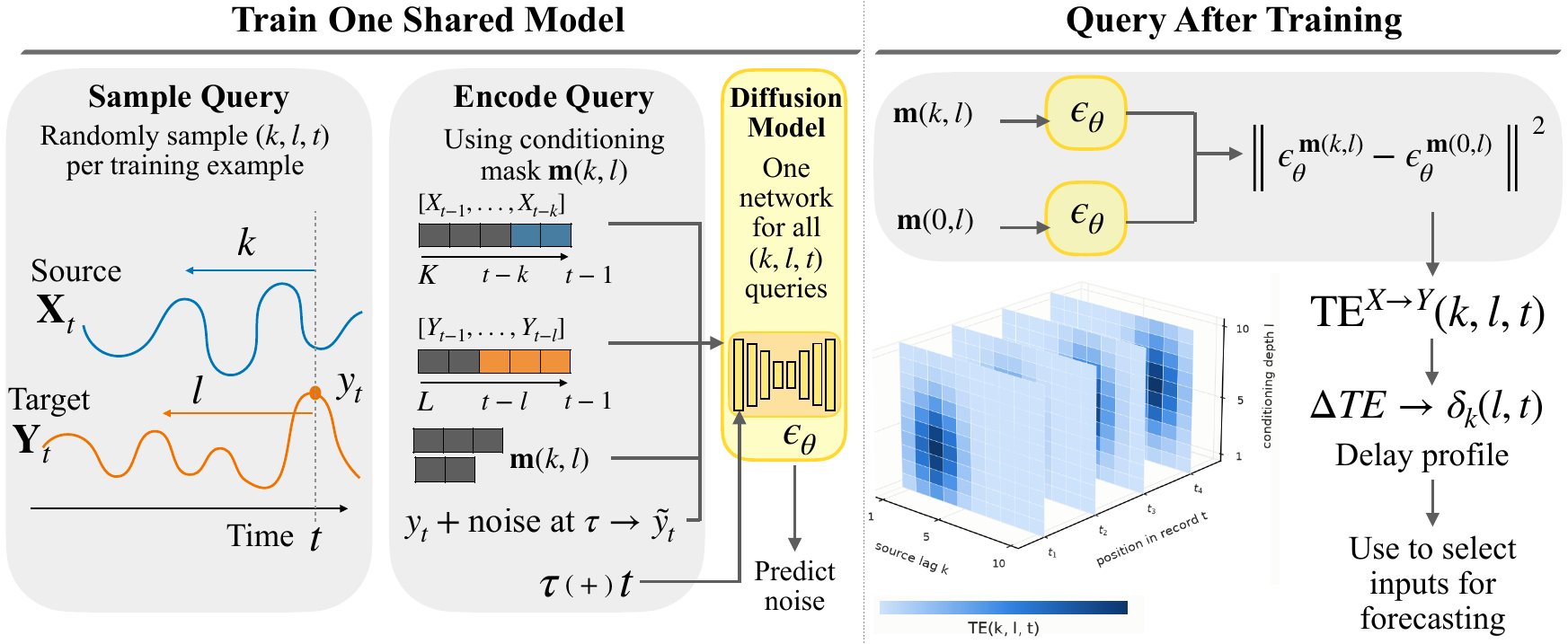}
\end{center}
\caption{Overview of ATEMPO: During training, source depth \(k\), conditioning depth \(l\) and time position \(t\) are varied across training samples. A mask \(\mathbf{m}(k,l)\) specifies which source and target lags are observed, while \(t\) and diffusion time \(\tau\) conditions through the same pathway. After training, the same network is evaluated with and without the source history; the squared difference between the resulting conditional scores estimates \(TE(k,l,t)\). Querying different \((k,l,t)\) recovers the TE surface without refitting, and differencing along \(k\) yields the time-varying delay profile. Using these lags as inputs to a forecasting model improves performance over consecutive-timestep baselines.}
\label{fig:contribution}
\end{figure}

Our key insight is that the estimators required across this surface differ in their conditioning information, not in the underlying estimation problem. We therefore condition a shared score model on source depth, target-conditioning depth, and temporal position, sampling these configurations during training. The resulting estimator, \textbf{ATEMPO} (Amortized Transfer Entropy Model Per-lag Over-time), learns the full TE surface in a single fit, turning quantities previously fixed during training into inference-time choices. 
Our contributions are:
\begin{contributionbox}
\begin{enumerate}[leftmargin=*, itemsep=1pt, topsep=2pt]

\item \textbf{Amortization over lags and over time.}
ATEMPO amortizes estimation across source depth, conditioning depth, and temporal position, replacing separate fits across the TE surface with one conditional score model. It achieves the lowest mean relative error \(8.3\pm3.3\%\) across synthetic systems in a single fit, whilst others require 18 fits. For time-varying systems, it maintains perfect lag-detection accuracy across all window lengths.

\item \textbf{Recovered delays provide useful inputs for downstream prediction.} 
Identifying where predictive information lies in a time series can provide more useful inputs than exposing a forecasting model to more history. On WEATHER-5K, delays selected using only 300 stations reduce iTransformer MSE by $7.25\%$ across all $5{,}672$ stations; using only half the benchmark's default input features still improves MSE by $4.6\%$. On three Neural Latent Benchmark tasks, adding ATEMPO's top five selected lags improves forecasting performance, achieving the highest or equal-best $R^2$. Averaged across tasks, ATEMPO improves $R^2$ by 4.1\% relative to the strongest competing baseline among eight methods.
%We demonstrate that identifying where predictive information lies in a time series can provide more useful inputs that simply exposing a forecasting model to more history.

\end{enumerate}
\end{contributionbox}
\section{Related Work}
\label{sec:related}

\paragraph{Lag selection with transfer entropy.} Choosing the lag badly is not a minor error: it inflates the estimate and produces spurious detections \citep{faes2011}. Much of the literature is therefore methods for identifying delayed dependence. Granger causality, which coincides with transfer entropy for linear Gaussian processes, is confined to linear couplings. 
Time-delayed mutual information \citep{fraser1986} handles nonlinear ones and recovers the direction of interaction where Granger causality fails \citep{li2018}, but cannot separate what the driving series contributes from what both series share through a common input. Conditioning on the target's own past resolves this, and is what transfer entropy does \citep{schreiber2000}. Built on it, momentary information transfer \citep{pompe2011} and $\mathrm{TE}_{\mathrm{SPO}}$ \citep{wibral2013} compare couplings across candidate delays, while non-uniform embedding~\citep{faes2011} admits lags one at a time by a greedy information criterion. Graph-based methods instead search dependencies over a maximum lag~\citep{runge2020,pamfil2020}. 
%These methods identify temporal dependencies; we instead estimate TE jointly across source and conditioning depths, allowing the contribution of each delay to be recovered from a shared TE surface.
A shared limitation is that when each candidate lag is estimated on its own, their errors are independent. This means they do not cancel when you difference TE estimates across lags, and that comparison is needed to identify the lag contributing the most information.

\paragraph{Time-varying coupling.} 
Time-resolved TE is typically obtained through local or sliding-window estimation, trading temporal resolution against the samples available within each window. 
% When dependence changes during a recording, time-resolved TE is conventionally estimated locally, for example through the sliding-window procedures such as those implemented in IDTxl~\citep{wollstadt2018idtxl} and commonly used in biomedical research~\citep{el2026reve, kontras2024core}. %CITE?
Related work on nonstationary causal discovery instead incorporates temporal information directly, for example CD-NOD~\citep{huang2020causal} introduces a time/domain index to represent changing distributions. ATEMPO combines these ideas differently: temporal position conditions a shared TE estimator trained over the complete recording, while a local window is used only to aggregate its estimates at evaluation. Window width is therefore decoupled from the samples available for fitting the score model.

\paragraph{Estimating transfer entropy.} Classical TE estimators inherit the difficulties of conditional mutual-information estimation. Nearest-neighbor methods need exponentially many samples as the conditioning set grows \citep{gao2015}. Variational lower bounds, built on generative models \citep{mondal2020}, discriminative critics \citep{belghazi2018mutual} or transformers \citep{luxembourg2025}, can become loose at large information values~\citep{mcallester2020}. Cross-entropy estimators escape that bound by taking a difference of entropies, as in CCMI \citep{mukherjee2020ccmi} and NJEE \citep{shalev2024}, but discretize continuous data to do it.
Diffusion-based estimators provide an alternative by expressing information quantities through differences of learned score functions. MINDE~\citep{franzese2024minde} introduced this approach for mutual information, and S$\Omega$I~\citep{bounoua2024somegai} showed that multiple score configurations required by a multivariate information quantity can be amortized within a single denoising model. TENDE \citep{munoz2025} applies conditional diffusion scores specifically to transfer entropy and is the closest estimator to ours. However, TENDE fixes the source and target history configuration and window size before training, so the cost remains one fit per hypothesis and per window.

No method therefore varies all three axes. ATEMPO instead amortizes the TE query itself: $(k,l)$ specify the conditioning set and $t$ the temporal position, allowing one model to recover $TE(k,l,t)$ without refitting.%Section~\ref{sec:background} gives the diffusion identity this rests on, 
Section~\ref{sec:problem} states the estimation problem and Section~\ref{sec:method} describes the model.
 
\section{Problem setting}
\label{sec:problem}
We seek to identify when information from a source \(X\) reaches a target \(Y\), after accounting for information already contained in the target's own past. 
To do this, we observe one recording $\{(x_t,y_t)\}_{t=1}^{T}$ of a bivariate process with $X_t\in\mathbb{R}^{N_x}$ the source and $Y_t\in\mathbb{R}^{N_y}$ the target. Write $\mathbf{X}^{(k)}_t:=(X_{t-1},\dots,X_{t-k})$ and $\mathbf{Y}^{(l)}_t:=(Y_{t-1},\dots,Y_{t-l})$ for histories of depth $k$ and $l$, empty at depth zero, and $K,L$ for the deepest considered. Transfer entropy \citep{schreiber2000} is the conditional mutual information
\begin{equation}
  \mathrm{TE}^{X\to Y}(k,l)\;:=\;I\!\left(Y_t;\ \mathbf{X}^{(k)}_t\ \middle|\ \mathbf{Y}^{(l)}_t\right)
  \;=\;\mathbb{E}\!\left[\log\frac{p\bigl(Y_t\mid\mathbf{X}^{(k)}_t,\mathbf{Y}^{(l)}_t\bigr)}{p\bigl(Y_t\mid\mathbf{Y}^{(l)}_t\bigr)}\right],
  \label{eq:te}
\end{equation}
zero exactly when the source history adds nothing beyond the target's own past. Exchanging the roles of the two series gives $\mathrm{TE}^{Y\to X}$. Conditional mutual information is symmetric in its first two arguments, so the asymmetry comes not from the measure but from which series supplies the history.

\paragraph{Locating the delay.} A single evaluation of Eq.~\ref{eq:te} scores the block $\mathbf{X}^{(k)}_t$ as a whole: it says the last $k$ source samples jointly inform the target, not which of them does. Differencing in $k$ resolves the block, since $\mathbf{X}^{(k)}_t=(\mathbf{X}^{(k-1)}_t,X_{t-k})$ and the chain rule gives, for $k\ge1$,
\begin{equation}
  \delta_k(l)\;:=\;\mathrm{TE}^{X\to Y}(k,l)-\mathrm{TE}^{X\to Y}(k-1,l)
  \;=\;I\!\left(Y_t;\ X_{t-k}\ \middle|\ \mathbf{X}^{(k-1)}_t,\mathbf{Y}^{(l)}_t\right)\;\ge\;0 .
  \label{eq:profile}
\end{equation}

The vector $\boldsymbol{\delta}(l)=(\delta_1(l),\dots,\delta_K(l))$ is the \emph{delay profile}: a large entry at $k$ means the source informs the target $k$ steps later, and a run of large entries means the influence persists beyond information provided by earlier lags. 
By differencing adjacent values, Eq.~\ref{eq:profile} decomposes $\mathrm{TE}^{X\to Y}(k,l)$ from a cumulative measure of information in the first \(k\) source lags to a profile over individual delays.
Neither scoring delays in isolation nor conditioning on all others has this property; Appendix~\ref{app:profile} compares them.

The delay profile resolves information across source lags, but it still assumes that two quantities are known. First, its value depends on the target-history depth \((l)\): too little history may attribute to the source information already contained in the target's past, whilst unnecessarily deep conditioning increases the dimension of the estimation problem. We therefore require the profile across both source depth \(k\) and conditioning depth \(l\). 
A long recording adds a third axis if we don't want to be constrained to lags fixed across time. We therefore seek not a single TE estimate, but a surface over source depth, conditioning depth and time: $\mathrm{TE}(k,l,t)$ for Eq.~\ref{eq:te} with the law taken over a window of width $w$ centred at $t$, and $\mathcal{I}$ for the admissible centres.
Recovering this surface yields the delay profile at every conditioning depth and position by differencing along its first axis.

\begin{problem}[Transfer entropy over lags and time]
\label{prob:main}
Given a single recording of two series, determine how much the past of each informs the present of the other at every combination of three quantities: the delay $k\le K$ at which the source is read, the depth $l\le L$ of the target's own past that is held fixed, and the position $t\in\mathcal{I}$ in the recording at which the estimate is taken. Formally, recover the surface
\begin{equation}
  \bigl\{\,\mathrm{TE}^{X\to Y}(k,l,t),\ \mathrm{TE}^{Y\to X}(k,l,t)\,\bigr\},
  \qquad
  (k,l,t)\in\{1,\dots,K\}\times\{1,\dots,L\}\times\mathcal{I},
  \label{eq:problem}
\end{equation}
without fitting a separate estimator for every point on that surface.
\end{problem}

\paragraph{Computational and statistical cost.} One estimator per cell of Eq.~\ref{eq:problem} costs $2(K{+}1)(L{+}1)\lvert\mathcal{I}\rvert$ fits. The conditioning set has dimension $kN_x+lN_y$, growing along the very axes being scanned, while a window of width $w$ supplies at most $w$ rows for every cell it contains. Widening the window buys samples at the cost of the time resolution the third axis exists to provide.

\section{Method}
\label{sec:method}

To estimate transfer entropy across source depth \(k\), conditioning depth \(l\) and position \(t\), existing estimators would require a separate fit at each cell, corresponding to every hypothesis about where and when information flows. ATEMPO avoids these repeated fits through a single principle: the quantities that specify the TE hypothesis are inputs to the estimator, rather than properties fixed during training.
Diffusion-based TE estimation is particularly suited to this construction. Unlike estimators that require discretization or explicit density estimation, it operates directly on continuous variables and accommodates nonlinear dependence. More importantly, it expresses transfer entropy through two score functions that differ only in their conditioning information. We exploit this structure by training one score model across conditioning configurations and temporal positions. Source-history depth and target-history depth are specified by a conditioning mask, while position enters through the same conditioning path as diffusion time. After training, changing \((k,l,t)\) changes the model's conditioning rather than its parameters. An overview of our method is shown in Figure~\ref{fig:contribution}.
\subsection{Transfer entropy from score functions}
\label{sec:scores}
The log-ratio in Eq.~\ref{eq:te} is already a Kullback--Leibler divergence between two conditional distributions of the target. Taking the inner expectation over $Y_t$ and the outer over the histories,  transfer entropy is expected divergence between these distributions, which differ only in whether the source history is supplied. ATEMPO is based on the principle that if this divergence can be estimated by a model whose conditioning can be varied, then \(k\) and \(l\) need not be fixed when the model is trained.

Diffusion-based estimators express the divergence without explicitly estimating the density, but instead through two score functions, obtained by regression. Let a variance-preserving SDE \citep{song2021score} with schedule $\beta(\tau)>0$ act on the target alone for $\tau\in[0,\bar{\tau}]$, giving $u_\tau=\kappa_\tau u_0+\sigma_\tau\varepsilon$ with $\kappa_\tau=\exp(-\tfrac12\int_0^\tau\!\beta(\tau))$, $\sigma_\tau^2=1-\kappa_\tau^2$ and $\varepsilon\sim\mathcal{N}(0,I)$, so the transition kernel is closed-form and no trajectory is simulated. The histories enter only as conditioning. Writing $p_\tau(\cdot\mid c)$ for the law of $u_\tau$ when $u_0\sim p(\cdot\mid c)$, \citet{franzese2024minde} give
\begin{equation}
  \mathbb{E}_{c}\bigl[D_{\mathrm{KL}}(p(\cdot\mid c)\,\|\,q(\cdot\mid c))\bigr]
  =\int_0^{\bar{\tau}}\frac{\beta(\tau)}{2}\,
  \mathbb{E}_{c}\,\mathbb{E}_{u\sim p_\tau(\cdot\mid c)}
  \!\left\lVert\nabla_u\log p_\tau(u\mid c)-\nabla_u\log q_\tau(u\mid c)\right\rVert^2
  \mathrm{d}\tau ,
  \label{eq:kl}
\end{equation}
up to a residual at $\tau=\bar\tau$ exponentially small in $\int_0^{\bar\tau}\beta(\tau)$, since both diffused laws converge to $\mathcal{N}(0,I)$. The conditional score is fixed by the posterior mean of the noise, $\nabla_u\log p_\tau(u\mid c)=-\,\mathbb{E}[\varepsilon\mid u_\tau=u,c]/\sigma_\tau$, which a network $\epsilon_\theta(u,\tau,c)$ attains at the optimum of the denoising objective \citep{vincent2011connection,ho2020denoising}. Substituting both scores into Eq.~\ref{eq:kl} and drawing $\tau$ from $q(\tau)\propto\beta(\tau)/\sigma_\tau^{2}$ with normaliser $Z=\int\beta\sigma^{-2}\mathrm{d}\tau$ gives
\begin{equation}
  \mathrm{TE}^{X\to Y}(k,l)\;=\;\frac{Z}{2}\;
  \mathbb{E}_{\tau\sim q,\ \varepsilon}
  \left\lVert \epsilon_\theta\bigl(u_\tau,\tau,(\mathbf{x},\mathbf{z})\bigr)-\epsilon_\theta\bigl(u_\tau,\tau,\mathbf{z}\bigr)\right\rVert^2 ,
  \label{eq:estimator}
\end{equation}
so one cell costs two forward passes under different conditioning. 
TENDE \citep{munoz2025} evaluates Eq.~\ref{eq:estimator} at a fixed $(k,l)$, with the conditioning structure built into the network's input layer before training. Therefore returning the full \(TE(k,l,t)\) surface would require a fit per cell.

Our key observation is that changing \((k,l)\) does not change the score-estimation problem, only which variables the score is conditioned on. We therefore make the conditioning structure itself an input to a shared score model.

\subsection{Amortizing over the transfer-entropy surface}
\label{sec:amortise}

ATEMPO trains one score network across the full \((k,l,t)\) surface. To vary \((k,l)\) without refitting, we provide the maximum histories and use a mask to encode which lags are conditioned on:
%ATEMPO trains one score network to represent the family of conditional laws required across \((k,l,t)\). We encode both forms of information explicitly and vary them during training through the conditioning argument \(c\). $c$ selects the law, so making $(k,l)$ an argument of the network rather than a property of the fit requires only that $c$ be able to name any cell. We replace the three-slot indicator of \citet{munoz2025} with one slot per lag,
\begin{equation}
  \mathbf{m}(k,l)\;=\;\bigl[m_y,\ m_{x,1},\dots,m_{x,K},\ m_{z,1},\dots,m_{z,L}\bigr]\in\{-1,0,1\}^{1+K+L},
  \label{eq:encoding}
\end{equation}
where $m_y=1$ marks the variable whose score is learned, $0$ a variable supplied as conditioning, and $-1$ a variable withheld.
For cell \((k,l)\), the first \(k\) source lags and \(l\) target lags are conditioned on and all deeper lags are withheld. Withheld values are zeroed before projection, while the mask $\mathbf{m}$ is also supplied to the network to distinguish them from observed zeros. Thus, changing \((k,l)\) changes only the mask, not the network.
%The cell $(k,l)$ is addressed by setting $m_{x,j}=0$ for $j\le k$ and $-1$ otherwise, and likewise $m_{z,j}$ at depth $l$. Withheld lags are zeroed before projection, and $\mathbf{m}$ is supplied alongside the data so a withheld lag is distinguishable from one observed to be zero. A single input representation can therefore address every \((k,l)\) cell, as changing these depths changes the mask and not the network.

\paragraph{Training objective.} Combining denoising score matching with the cell drawn per example,
\begin{lossbox}
% \begin{equation}
%   \mathcal{L}(\theta)\;=\;
%   \mathbb{E}_{\substack{(k,l)\sim\mathcal{U},\ t\sim\mathcal{I}\\ \tau\sim q,\ \varepsilon\sim\mathcal{N}(0,I)}}
%   \left\lVert\,
%   \epsilon_\theta\bigl(\kappa_\tau y_t+\sigma_\tau\varepsilon,\ \tau,\ t,\ \mathbf{m}(k,l),\ \mathbf{X}^{(K)}_t,\mathbf{Y}^{(L)}_t\bigr)-\varepsilon
%   \,\right\rVert^{2}
%   \label{eq:loss}
% \end{equation}
\begin{equation}
  \mathcal{L}(\theta)\;=\;
  \mathbb{E}_{\substack{(k,l)\sim\mathcal{U},\ t\sim\mathcal{I}\\ \tau\sim q,\ \varepsilon\sim\mathcal{N}(0,I)}}
  \Bigl\lVert\,
  \epsilon_\theta\bigl(
    \underbrace{\kappa_\tau y_t+\sigma_\tau\varepsilon,\ \tau}_{\substack{\text{noised target,}\\ \text{diffusion time}}}
    ,\ \underbrace{t,\ \mathbf{m}(k,l)}_{\substack{\text{sampled time,}\\ \text{mask of sampled depths}}}
    ,\ \underbrace{\mathbf{X}^{(K)}_t,\mathbf{Y}^{(L)}_t}_{\substack{\text{maximum histories,}\\ \text{masked by }\mathbf{m}}}
  \bigr)-\varepsilon
  \,\Bigr\rVert^{2}
  \label{eq:loss}
\end{equation}
\end{lossbox}
\noindent with $(k,l)\sim\mathcal{U}\{0,\dots,K\}\times\mathcal{U}\{0,\dots,L\}$. Eq.~\ref{eq:loss} differs from the objective of \citet{munoz2025} in that: $(k,l)$ is drawn per example instead of fixed before training, and the position $t$ is supplied as a conditioning argument. The forward process, the denoising target and the estimator are unchanged. That single change replaces the $(K{+}1)(L{+}1)$ fits per window of Eq.~\ref{eq:problem} with one fit in total.
d
\paragraph{Conditioning on temporal position.}
The mask amortizes over $(k,l)$, but detecting a lag that changes over the recording additionally requires the score to depend on position $t$. Rather than fitting a separate model in each time window, we condition the shared score network on $t$.
Diffusion models already condition on diffusion time $\tau$, which is sinusoidally embedded and injected into every residual block as a scale and shift. We reuse this pathway for recording position, normalizing $t$ to $[0,1]$ and embedding it on the same basis with a separate frequency scale $s_t$. The embeddings are summed, $h=h_\tau+h_t$, before the modulation 
\begin{equation}
  h^{(b+1)}\;=\;h^{(b)}+f_b\!\left(\mathrm{GroupNorm}(h^{(b)})\odot(1+\gamma_b)+\beta_b\right),
  \qquad (\gamma_b,\beta_b)=W_b h .
  \label{eq:film}
\end{equation}
Routing \(t\) through this pathway lets position information, like diffusion time, modulate features throughout the network rather than enter only at the input projection. 
Thus $\tau$ identifies where in the diffusion process the sample lies, while $t$ tells it where it lies in the recording.
%This path is multiplicative, which is why $t$ is routed through it rather than appended to the input: the gain $1+\gamma_b$ scales a feature's response, whereas a scalar entering once at the input projection contributes an additive shift the network can learn to ignore. Since $\tau$ is drawn per row, a mini-batch spans many noise levels at once, so the network cannot specialize to one and must read both signals at every pass.

After one fit, any point on the TE surface is recovered by Eq.~\ref{eq:estimator} with the two conditionings addressed by the encoding,
\begin{equation}
  \mathrm{TE}^{X\to Y}(k,l,t)\;=\;\frac{Z}{2}\;
  \mathbb{E}_{\tau\sim q,\ (y,\mathbf{x},\mathbf{z})\in\mathcal{W}_t,\ \varepsilon}
  \left\lVert \epsilon_\theta^{\,\mathbf{m}(k,l)}-\epsilon_\theta^{\,\mathbf{m}(0,l)}\right\rVert^2 ,
  \label{eq:final}
\end{equation}
where $\mathbf{m}(0,l)$ denotes that every source lag is withheld, and the superscript abbreviates the arguments of Eq.~\ref{eq:loss}.

The position $t$ conditions the score network, whereas the window $\mathcal W_t$ only determines which local estimates are averaged. We therefore choose its width at evaluation rather than fixing it during training.
This temporal sharing assumes that variation in the conditional distribution across the recording can be represented by a shared function of the observations and position \(t\), rather than requiring each local window to define an independent estimation problem.
The delay profile Eq.~\ref{eq:profile} then follows by differencing Eq.~\ref{eq:final} in $k$. Thus, ATEMPO separates training from the hypothesis required for delay discovery: one network learns the conditional scores, the TE surface is queried after training, and differencing that surface locates when source information reaches the target.

\section{Experiments}
\label{sec:datasets}
%% Weather 5k
\paragraph{Datasets.}
WEATHER-5K~\citep{weather5k} contains hourly observations of five meteorological variables from 5,672 stations over 2014-2023. We use the benchmark train/validation/test split (2014-2021/2022/2023) and evaluate forecasting on all 5,672 stations and a fixed 300-station subset. ATEMPO is fit on the 300 station's training data, once for each of the 25 ordered variable pairs (\(K=336\)h, \(L=24\)h), and the resulting lags are shared across stations. This tests whether delay structure learnt from a small subset remains useful at full-dataset scale. We test three types of downstream forecaster models on WEATHER5K, transformers that naively support non-consecutive inputs (iTransformer, DLinear), transformers that do not (PatchTST, Pyraformer) and zero-shot pretrained foundation model, Chronos-2. In Appendix~\ref{sec:appendix-additional-models} we present full results passing only the selected subset of 48 input rows, and for Chronos-2 passing the full 336-hour window with non-selected values marked as missing.

We also examine whether the top 5 ATEMPO-selected lags, chosen on training data only, can improve forecasting on held-out data in three Brain-Computer interface tasks from the Neural Latents Benchmark (NLB)~\citep{PeiYe2021NeuralLatents}.
We evaluate on the motor cortical (MC) tasks which involve reaching for a target or maze, establishing physiologically delays that provide a basis for evaluating lag selection.
For a randomly changing target, NLB velocity aims to predict the finger velocity 200ms ahead using its past velocity and selected lags of the target-position error. NLB decoding uses the same task but predicts current finger velocity from neural spiking activity, comparing lag selection against the benchmark's standard assumption of a fixed 140ms neural lead. MC\_Maze predict hand velocity for a maze task at movement onset using only preparatory neural activity, making accurate neural lead-time selection essential. 
All methods use the same ridge regression forecaster and are evaluated using held-out $R^2$.

To compare baselines for exact TE estimates and lag detection, we use seven synthetic datasets with ground truth TE values; an additional five time-varying datasets makes 12 with ground truth lags.  Experiments across window sizes are performed on three of the datasets where the source-to-target delay switches from lag at \(k=2\) to \(k=5\) halfway through an 8,000 sample recording. These three datasets cover linear, mean-blind squared and effect-size-matched squared coupling. Further dataset details are in Appendix~\ref{sec:appendix-dataset-details}.
To extend beyond time series data, we use MMTT-Bench~\citep{mmtt-bench}, which pairs a synthetic signal with natural language descriptions of what the signal is about to do. Full experimental protocols are detailed in Appendix~\ref{sec:appendix-dataset-implementations}.
%To extend beyond fully generated time series data, we use MMTT-Bench~\citep{mmtt-bench}, which pairs a synthetic signal with natural language descriptions of what the signal is about to do.  Further information about our synthetic dataset can be found in Appendix~\ref{sec:appendix-dataset-details}.
%
\paragraph{Baselines.}
We compare our method to eight baselines, all of which require a model trained for each (k,l) pair: TENDE~\citep{munoz2025} uses a diffusion architecture; 
NPEET computes CMI from kNN statistics; neural estimators NJEE and MINE  calculate TE as the difference between entropies; CCMI estimates CMI directly using a binary classifier; and PCMI computes CMI on symbolic encodings of the values rather than the values themselves.
To extend our comparison for lag detection we include baselines Liang-Kleeman~\citep{liang2013liang} and DTW, which do not estimate transfer entropy in nats directly, and are therefore excluded from evaluations of explicit TE values. 
%Both estimators are reported as a cumulative sum over source lags \((1..k)\) so that \(\Delta TE\) recovers exactly the per-lag quantity each method computes.
Full implementation details can be found in Appendix~\ref{sec:appendix-baseline-implementations}.

\subsection{Results}
\label{sec:results}
\begin{table}[t]
\scriptsize
\centering
\caption{Forecasting error on WEATHER-5K at the benchmark's input size for two forecasters: iTransformer and DLinear. MSE is averaged over the five variables and over training seeds, with standard deviation across seeds \(<0.002\) for every run.}
\label{tab:weather5k}
\begin{tabular}{lcc|cc|c}
\toprule
& \multicolumn{2}{c|}{iTransformer} & \multicolumn{2}{c|}{DLinear} & \\
& MSE ($\downarrow$)& $\Delta$ vs default ($\downarrow$) & MSE ($\downarrow$)& $\Delta$ vs default ($\downarrow$) & \makecell{selection\\cost (h)} \\
\midrule
\multicolumn{6}{l}{\textit{(a) all 5{,}672 stations}}\\
Default & 0.5562 & - & 0.5840 & - & 0 \\
ATEMPO & \bf{0.5158} & \cellcolor{green!25}-7.25\% & \bf{0.5252} & \cellcolor{green!25}-10.07\%& 1.8 \\
% \cmidrule(l){1-6}
% \textit{Full window}$^\dagger$ & 0.5133 & -7.70\% & -- & -- & - \\
\midrule
\multicolumn{6}{l}{\textit{(b) 300 stations}}\\
Default & 0.5223 & - & 0.5510 & - & 0 \\
Random & 0.5066 & -3.00\% & 0.5040 & -8.53\% & 0 \\
Liang & 0.5039 & -3.51\% & 0.5244 & -4.83\% & 0.5 \\
DTW & 0.5021 & -3.85\% & 0.5198& -5.66\% & 0.2 \\
NPEET & 0.5005 & -4.16\% & 0.5192 & -5.77\% & 2.6 \\
PCMI & 0.4994 & -4.38\% & 0.4979 & -9.64\% & 1.3 \\
MINE & 0.5222 & -0.01\% & 0.5147 & -6.58\% & 4.6 \\
NJEE & 0.4971 & -4.81\% & 0.4965 & -9.89\% & 2.9 \\
CCMI & \bf{0.4904}& \cellcolor{green!25}-6.10\% & \bf{0.4917} & \cellcolor{green!25}-10.77\% & 4.4 \\
ATEMPO & \bf{0.4904} & \cellcolor{green!25}-6.10\% & 0.4948 & -10.21\% & 1.8 \\
% \cmidrule(l){1-6}
% \textit{Full window}$^\dagger$ & 0.4918& -5.84\% & 0.4821& -12.50\% & - \\
\bottomrule
\end{tabular}
% \par\smallskip\footnotesize $^\dagger$ No lag is selected, so there is no selection cost.
\end{table}
\begin{figure}[h]
\begin{center}
\includegraphics[width=\textwidth]{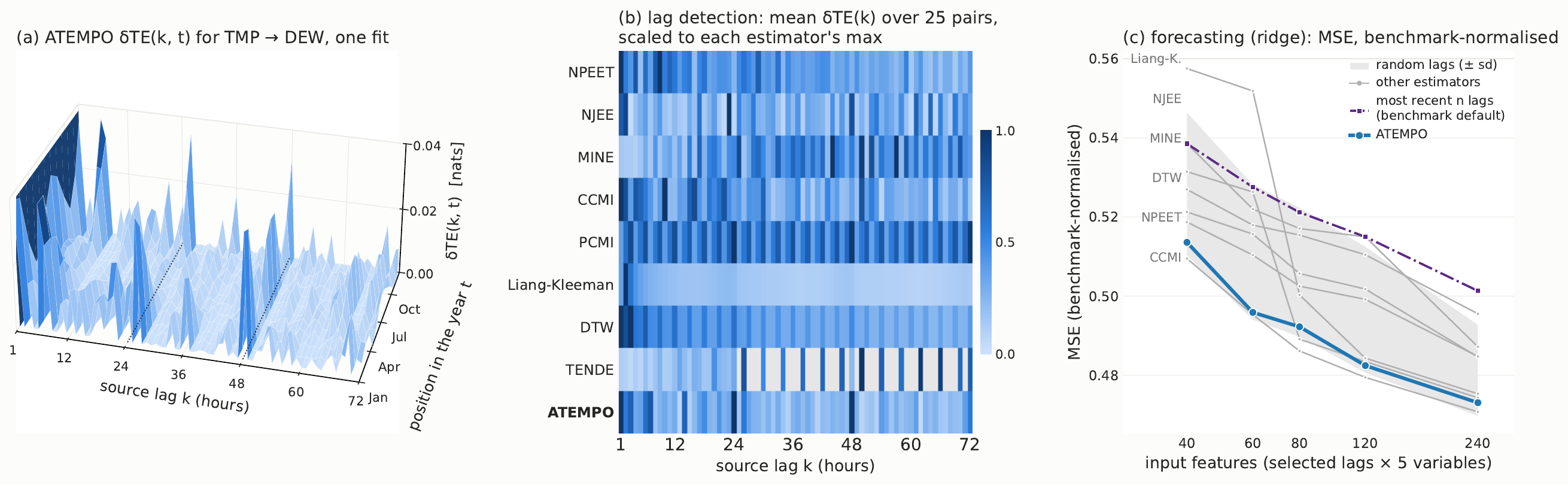}
\end{center}
\caption{ATEMPO applied to WEATHER5K. a) A delay profile between source variable temperature (TMP) and target variable dew point temperature (DEW) over time, pooled over 300 stations b) The mean delay profile \(\delta_k = TE(k) - TE(k-1)\) averaged over 25 variable pairs for each baseline. The darker colors are the lags carrying the most information. c) Forecasting performance (MSE) from using each method's top ranked \(n\) features as inputs.}
\label{fig:w5k}
\end{figure}
\begin{table}[t]
\centering
\scriptsize
\caption{Mean relative \% error in TE estimation, lag detection accuracy and lag selection forecasting performance ($R^2$) across synthetic and real datasets.}
\label{tab:te-estimate}
\begin{tabular}{lc|cc|cccc}
\toprule
 & TE error \% $(\downarrow)$ & \multicolumn{2}{c|}{lag detection: acc. $(\uparrow)$} & \multicolumn{4}{c}{lag selection: $R^2$, 5 lags $(\uparrow)$} \\
 & \makecell{synthetic\\datasets} & \makecell{synthetic\\datasets} & \makecell{MMTT-\\Bench}  & \makecell{NLB\\velocity} & \makecell{NLB\\decoding} & \makecell{MC\_Maze\\decoding} &
 \makecell{Mean\\$R^2$} \\
\midrule
\makecell{Liang--\\Kleeman} & - & 0.46 $\pm$ 0.34 & 0.00  & 0.789 & 0.458 & 0.582 & 0.610\\
DTW & - & 0.71 $\pm$ 0.42 & 0.00   & 0.794 & 0.402 & 0.554 & 0.583\\
NPEET & 56.0 $\pm$ 38.7 & 0.91 $\pm$ 0.24 & 0.33  & 0.805 & 0.444 & 0.504 & 0.584\\
PCMI & 55.1 $\pm$ 22.1 & 0.92 $\pm$ 0.19 & \textbf{1.00}  & 0.794 & 0.456 & 0.486 & 0.579\\
NJEE & 32.9 $\pm$ 13.1 & 0.92 $\pm$ 0.13 & 0.72 & 0.809 & 0.402 & 0.504 & 0.572\\
CCMI & 24.0 $\pm$ 26.4 & 0.95 $\pm$ 0.08 & 0.78   & \textbf{0.813}$^{*}$ & 0.417 & 0.592 & 0.607\\
TENDE & 16.3 $\pm$ 14.0 & 0.95 $\pm$ 0.12 & 0.94 & 0.809 & 0.432 & 0.504 & 0.582\\
MINE & 11.9 $\pm$ 8.3 & 0.97 $\pm$ 0.08 & 0.83  & 0.785 & 0.412 & 0.603 & 0.600\\
ATEMPO & \textbf{8.3 $\pm$ 3.3} & \textbf{0.98 $\pm$ 0.07} & 0.94  & \textbf{0.812}$^{*}$ & \textbf{0.474} & \textbf{0.619} & \textbf{0.635}\\
\bottomrule
\end{tabular}
\par\smallskip\scriptsize $^{*}$Standard error $= 0.003$ for paired differences against ATEMPO in NLB velocity task
\end{table}
\paragraph{WEATHER-5K}
In a single fit, ATEMPO ranks \(n\) features from the preceding 336 hours. Despite selecting these lags using only 300 stations, the resulting inputs reduce forecasting MSE by 7.25\% relative to the WEATHER5K default across all 5,672 stations. This shows that the selected lag structure transfers beyond the stations on which it was estimated. The selected lags also exhibit structure consistent with known meteorological periodicities: prominent delays occur at 24-hour intervals, as expected from the diurnal cycle~\citep{dai2004diurnal, wu2021autoformer} and further examined in Appendix~\ref{sec:appendix-weather5k-results}.
On the 300-station subset, Table~\ref{tab:weather5k} shows that ATEMPO-selected inputs reduce forecasting MSE by 6.10\% using 240 features and 4.6\% using only 120, half the features used by benchmark default we are comparing with. This ranking of ATEMPO-selected lags beating both random and default inputs holds across 4/5 tested forecaster models. CCMI is the only baseline to achieve comparable forecasting performance, but cannot amortize estimation across lags: with conditioning depth fixed, evaluating \(K=336\) source lags for each of the 25 ordered variable pairs requires 8,400 CCMI fits, compared with ATEMPO's 25. Therefore lag selection costs 4.4 hours for CCMI,  1.8 for ATEMPO.
% MINIPAGE TOP HERE
% TO DO CAPTIONS
\begin{figure}[t]
    \centering
  \begin{minipage}[t][6.3cm][c]{0.48\textwidth}
    \centering
    \includegraphics[width=\textwidth]{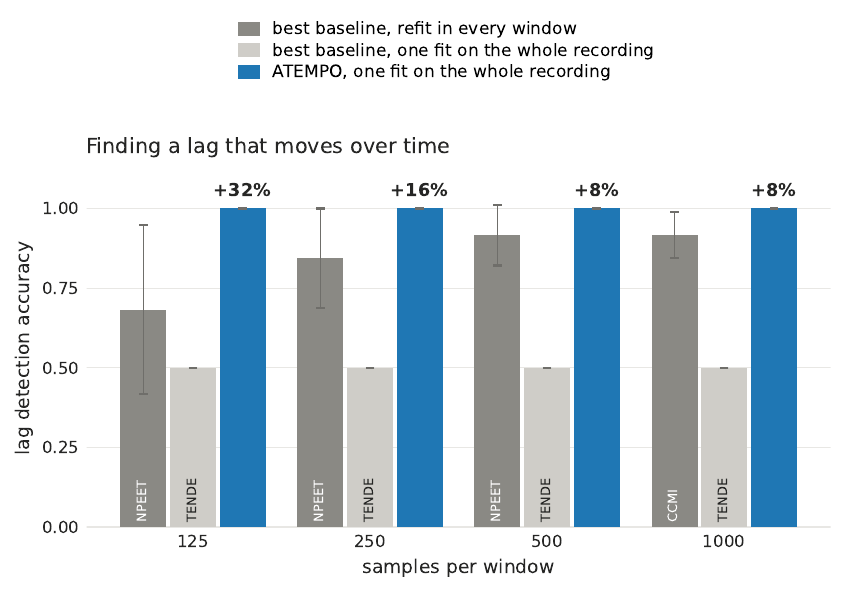}
    \captionof{figure}{Lag detection accuracy against window length, mean and std over three time-varying systems. Each method reports one lag per window; accuracy is the fraction of windows whose top-1 lag is the true one.}
    \label{fig:ablation}
    \end{minipage}
  \hfill % Adds horizontal stretching space between the minipages
  % Second minipage for the right table
   \begin{minipage}[t][6.3cm][c]{0.48\textwidth}
    \centering
    \includegraphics[width=\textwidth]{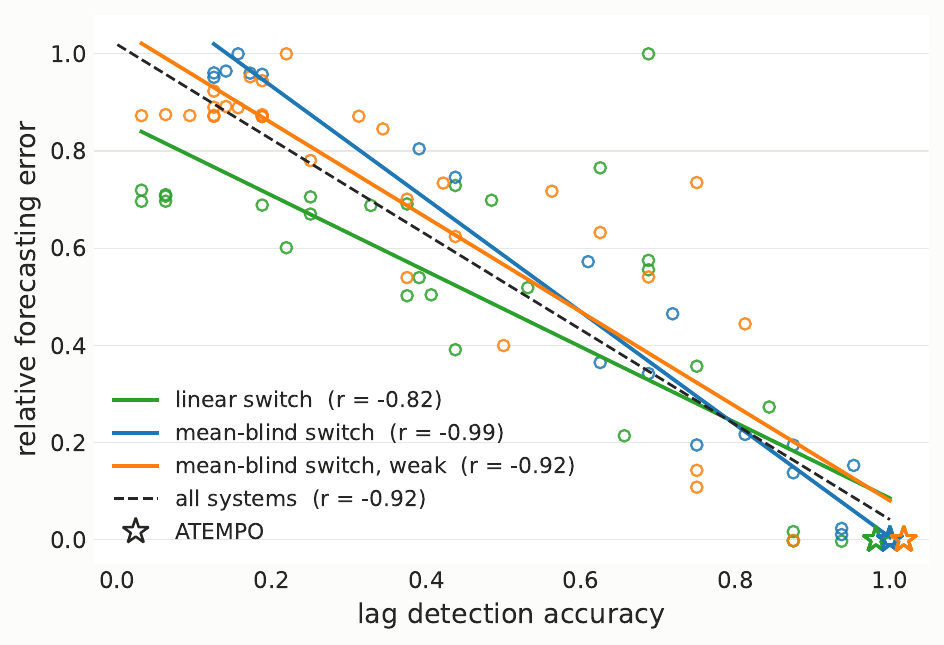}
    \captionof{figure}{Correlation between lag detection accuracy and forecasting error for three time-varying systems. Relative forecasting error is the MSE error compared to the MSE when using true lags for forecasting.}
    \label{fig:detection-forcasting}
    \end{minipage}
\end{figure}
\paragraph{Neural Latent Benchmark}
Each method selects five lags on the training part of three MC tasks from Neural Latents Benchmark~\cite{PeiYe2021NeuralLatents}. Results in Table~\ref{tab:te-estimate} show forecasting performance on held-out data when 5 selected lags are fed to the same ridge forecaster (paired bootstrap
standard errors $\le 0.003,0.010,0.011$ for each task).
ATEMPO's selections result in the best or equal performance, resulting in a mean $R^2$ 4.1\% higher than Liang-Kleeman, the next closest baseline. Appendix~\ref{sec:appendix-neural} shows results at different lag budgets, with ATEMPO performing best compared with three further methods: most recent lags, cross-correlation and the benchmark's fixed lag value.
\paragraph{Synthetic Datasets}
On synthetic datasets with ground truth TE and known lags, Table~\ref{tab:te-estimate} shows ATEMPO produces the most accurate transfer entropy estimates and the best lag detection accuracy. It achieves this in a single fit for source depths \(K=6\) and target depths \(L=3\), where other estimators require \(K\times L=\) 18 separate fits, one for each (k,l) pair. Comparing to the most similar model architecture TENDE, the time per fit is approximately 5.8 minutes for both TENDE and ATEMPO, so amortizing over the lag grid costs nothing in terms of time-per-fit, but produces a cost saving proportional to \(K\times L\) overall. Figure~\ref{fig:ablation} demonstrates the value of conditioning on time position on three synthetic datasets where the source-to-target delay changes partway through a recording. Without time conditioning an estimator has two choices. It can refit every window, shown in dark gray, which works while windows are long but decays as they shrink as each fit sees fewer samples - the best baseline estimator falls from 0.92 to 0.68 lag detection accuracy as window size drops from 1000 to 125. Alternatively it can fit once on the whole recording and report a single lag, shown in light gray, but this cannot exceed 0.5 accuracy when the lag switches once within the dataset. ATEMPO is fitted once on the same 8,000 samples and recovers the lag in every window at every length. The margin over the best refit baseline grows from 8-32\% as windows shrink.

Using the same three datasets, we quantify the extent to which accurate lag detection improves downstream forecasting. Figure~\ref{fig:detection-forcasting} shows the correlation between detection rate and MSE improvement, where each point is one estimator at one window length. A least-squares fit  per dataset gives the Pearson correlations \(r = [-0.82,-0.99,-0.92]\) per system (\(n=36\), two-sided t-test on r gives nominal p between \(10^{-9}\) and \(10^{-31}\)). Full results can be found in Appendix~\ref{sec:appendix-synthetic-results}.

\paragraph{MMTT-Bench}
The value of utilizing information driving time series is not restricted to a single modality. 
MMTT-Bench~\citep{mmtt-bench} pairs a synthetic signal with natural language descriptions of what the signal is about to do. Natively the text description at step \(t\) predicts the time series at \(t+1\), a lag of one. Re-aligning the two streams plants an arbitrary delay \(m\), giving a controlled ground truth in a multimodal forecasting setting. The benchmark supplies correct and irrelevant descriptions per timestep, where irrelevant text adds no additional information to the time series forecasting task.
Our estimator recovers the planted lag perfectly for five out of six lags and identified zero lags in the null irrelevant category, equaling TENDE and beating all other baselines except for PCMI in comparable processing time. Furthermore ATEMPO correctly estimates TE values highest for the correct text compared to irrelevant (0.52 to 0.14 nats), and this corresponds to an improvement in forecasting MSE of 41\% for correct. Full results are available in Appendix~\ref{sec:appendix-mmtt-results}.
\begin{table}[t]
\centering
\small
\caption{Lag detection and forecasting under a time-varying lag. Results are averaged over three synthetic systems. Lag detection accuracy is the fraction of evaluation windows in which the top-ranked lag equals the true lag. Relative forecasting error is the MSE error compared to the MSE when using true lags for forecasting. Per-window methods are refit independently using only the samples in each window. Fit-once methods are trained on the complete 8,000 sample recording. Only time-conditioned ATEMPO can vary its estimated lag across windows without refitting.}
\label{tab:ablation}
\resizebox{0.87\linewidth}{!}{%
\begin{tabular}{llcccc|cccc}
\toprule
& & \multicolumn{4}{c}{lag detection accuracy ($\uparrow$)} & \multicolumn{4}{c}{Relative forecasting error ($\downarrow$)}\\
& samples per window & 125 & 250 & 500 & 1000 & 125 & 250 & 500 & 1000 \\
\midrule
\multicolumn{10}{l}{\textit{fitted per window}} \\
& NPEET & 0.68 & 0.84 & 0.92 & 0.92 & 1.83 & 1.09 & 1.05 & 1.01 \\
& NJEE & 0.35 & 0.44 & 0.62 & 0.88 & 2.80 & 1.83 & 1.17 & 1.09 \\
& MINE & 0.32 & 0.52 & 0.67 & 0.92 & 3.87 & 1.28 & 1.17 & 1.02 \\
& CCMI & 0.19 & 0.42 & 0.54 & 0.92 & 4.70 & 2.10 & 1.16 & \textbf{1.00} \\
& PCMI & 0.29 & 0.48 & 0.67 & 0.75 & 4.98 & 2.21 & 1.21 & 1.08 \\
& Liang-Kleeman & 0.23 & 0.20 & 0.23 & 0.25 & 5.69 & 5.68 & 5.64 & 5.70 \\
& DTW & 0.43 & 0.48 & 0.73 & 0.71 & 2.90 & 3.35 & 1.11 & 1.13 \\
& TENDE & 0.11 & 0.11 & 0.35 & 0.88 & 5.73 & 5.90 & 2.10 & 1.07 \\
& ATEMPO (no time conditioning) & 0.06 & 0.12 & 0.62 & \textbf{1.00} & 5.73 & 5.72 & 1.91 & \textbf{1.00} \\
\midrule
\multicolumn{10}{l}{\textit{fitted once on the whole recording}} \\
& NPEET & 0.50 & 0.50 & 0.50 & 0.50 & 4.50 & 4.50 & 4.50 & 4.50 \\
& NJEE & 0.50 & 0.50 & 0.50 & 0.50 & 4.00 & 4.00 & 4.00 & 4.00 \\
& MINE & 0.50 & 0.50 & 0.50 & 0.50 & 4.01 & 4.01 & 4.01 & 4.01 \\
& CCMI & 0.50 & 0.50 & 0.50 & 0.50 & 4.00 & 4.00 & 4.00 & 4.00 \\
& PCMI & 0.50 & 0.50 & 0.50 & 0.50 & 4.00 & 4.00 & 4.00 & 4.00 \\
& Liang-Kleeman & 0.17 & 0.17 & 0.17 & 0.17 & 5.71 & 5.71 & 5.71 & 5.71 \\
& DTW & 0.50 & 0.50 & 0.50 & 0.50 & 4.49 & 4.49 & 4.49 & 4.49 \\
& TENDE & 0.50 & 0.50 & 0.50 & 0.50 & 4.01 & 4.01 & 4.01 & 4.01 \\
& ATEMPO (no time conditioning) & 0.64 & 0.60 & 0.60 & 0.58 & 3.96 & 4.01 & 4.00 & 4.01 \\
& \textbf{ATEMPO (time-conditioned)} & \textbf{1.00} & \textbf{1.00} & \textbf{1.00} & \textbf{1.00} & \textbf{1.00} & \textbf{1.00} & \textbf{1.00} & \textbf{1.00} \\
\bottomrule
\end{tabular}}
\end{table}
\paragraph{Ablation.}
Table~\ref{tab:ablation} separates two benefits of ATEMPO: amortizing over lag and conditioning on temporal position. When both models are fit once to the complete recording, amortizing over lag alone detects the moving lag in only 0.58-0.64 of windows. Adding temporal conditioning raises detection to 1.00 at every evaluation-window length. Thus, access to the full recording alone is insufficient: the estimator must also represent how the dependence changes over time.

Fitting estimators independently within each window has the opposite limitation. These models can represent local dependence, but their accuracy deteriorates as fewer samples are available for each fit. NPEET, the strongest baseline at 125 samples, falls from 0.92 detection at 1,000 samples to 0.68 at 125; ATEMPO without time conditioning falls from 1.00 to 0.06. Full ATEMPO avoids this trade-off: its time-conditioned score model is trained once on the complete recording, while the evaluation window determines only where TE estimates are aggregated. It therefore retains the statistical strength of the full recording while resolving local changes in lag, maintaining 1.00 detection with 125-sample evaluation windows. 

These differences propagate directly to forecasting. We report MSE relative to a forecaster given the known lag in every window, so 1.0 indicates oracle-level performance. As per-window lag detection deteriorates, forecasting error generally increases. Conversely, estimators fit once without temporal conditioning select a global lag and incur approximately \(4\times\) the oracle error despite using the entire recording. Full ATEMPO recovers the local lag in every window and consequently matches the oracle forecasting performance at every window length. 
\paragraph{Limitations.} ATEMPO amortizes transfer entropy estimation across lag, conditioning depth and time, but does not remove the statistical requirements of estimating conditional dependence. In particular, at large dimensionality, specified by \((K,l)\), there is a minimum requirement for data samples and model capacity. A second limitation concerns the recovered delay profile, as errors in the transfer entropy estimation can propagate through. Detailed analysis in Appendix~\ref{sec:appendix-physical-analysis} shows that not all selected lags add equal value to downstream forecasting, implying residual errors in TE estimation can lead to less-than-optimal lag selection. 
Finally the scope of our empirical evaluation is necessarily limited. Exact transfer entropy is available only for controlled processes, so quantitative estimation error is evaluated on synthetic systems which are approximately Gaussian and contain a single informative dimension. We complement these experiments with MMTT-Bench and WEATHER5K, where ground truth TE is unavailable and utility is instead assessed through known alignment or downstream forecasting. 

\paragraph{Future Work.}
The central idea of ATEMPO, turning quantities normally fixed during estimation into conditioning variables of a shared model, could be extended beyond pairwise transfer entropy to condition jointly on multiple sources. This would allow the estimator to distinguish information that is unique to one variable from information that is redundant or synergistic across variables, turning the present lag-selection procedure to joint variable-and-lag selection for high-dimensional forecasting. 
%The time-conditioned estimator also opens possibilities beyond discrete lag-selection setting studied here. Because one fit can be queried locally throughout a recording and at different evaluation-window widths, it provides a representation of how directional dependence evolves over time. Future work could use this surface for change-point detection, regime discovery or detecting changes in interaction delays, rather than reducing it to the strongest lag alone. 
%\section{Conclusion}
%\label{sec:conclusion}

\section{Conclusion}
Fixed input widths conflate two questions: how much history to expose to a model, and where within that history useful information actually lies. We introduced ATEMPO to separate them. By expressing transfer entropy through conditional diffusion scores and making lag, conditioning depth and temporal position inputs to a shared estimator, ATEMPO replaces separately fitted models with a single amortized fit. The resulting \(TE(k,l,t)\) surface can be queried across histories and time, and differenced along \(k\) to recover a time-varying delay profile. We show ATEMPO achieves the best mean TE estimations and lag-detection accuracy compared to 8 baselines. These recovered delays then reduce forecasting MSE by 5-7\% across the WEATHER5K dataset and improve $R^2$ by 4.1\% on average across three Neural Latent Benchmark tasks.

\clearpage
\newpage
\section*{AI use statement}
In this work, we used generative AI tools to check our implementations of TE estimator baselines which do not provide open-source code, and review our synthetic dataset generation. We did not use AI to help with any of the mathematical claims or proofs, propose or refine hypothesis or interpret results.

Additionally, we used generative AI to create some of the diagrams used in Figure 1 and generate the code to improve formatting in Figure 2. We also used it clean up and add comments to our code base to make it usable for others, and provide feedback on our research paper draft, specifically in terms of clarity and sentence structure. We have reviewed all AI-assisted work, all manuscript suggestions were made offline and only humans made changes directly to the text after AI feedback. All LLM-generated code was verified and tested for correctness using automated tests, and an author reviewed every AI-generated doc-string. We take full responsibility for the final content of this work. 

\section*{Ethics statement}
%(This section is \textbf{recommended} and does not count toward the page limit.)
% If authors feel that their paper submission raises questions regarding the Code
% of Ethics, they are encouraged to include a paragraph of Ethics Statement (at
% the end of the main text before references) to address potential concerns where
% appropriate. Topics include, but are not limited to, studies that involve human
% subjects, practices to data set releases, potentially harmful insights,
% methodologies and applications, potential conflicts of interest and sponsorship,
% discrimination/bias/fairness concerns, privacy and security issues, legal
% compliance, and research integrity issues (e.g., IRB, documentation, research
% ethics). This statement should not be more than 1 page.
This work develops a general-purpose method for estimating transfer entropy and identifying time-varying delayed dependencies in sequential data. Our experiments use synthetic data and existing public benchmarks.
The NLB dataset does include data derived from animals. It was collected under procedures and experiments that were approved by the Institutional Animal Care and Use Committees at the respective institutions. More details can be found in their original manuscript~\citep{PeiYe2021NeuralLatents}. Advances in predicting brain activity may be used to improve brain-computer interfaces, which can lead to better therapeutic devices for brain injury or disease. However it is also important to consider the potential negative impacts of this technology.

The potential benefits and risks of ATEMPO depend on its application and on the data to which it is applied. While identifying informative delays may improve forecasting and provide insight into temporal dependencies, the same methodology could be applied to sensitive data. Users should consider appropriate privacy, fairness and domain-specific requirements in such settings. We do not advocate using transfer entropy estimates alone for decision-making in high-stakes applications, instead we recommend human-in-the-loop to mitigate risks in these settings. 

\subsection*{Reproducibility statement}

Our ATEMPO method, all experiments and all figure and table generation is fully reproducible and made available on GitHub\footnote{\code}. The README contains clear instructions on how to reproduce all experiments, with one script per dataset. A notebook is available to generate all plots and figures from the saved results. Additionally, we have created a `run\_on\_your\_data.py` to make it easier for others to try ATEMPO. We use fully open-source datasets, with pre-processing details in Section~\ref{sec:datasets} and links to the data on GitHub. More details about our experimental protocol and baseline estimator implementations can be found in Appendix~\ref{sec:appendix-dataset-implementations} and~\ref{sec:appendix-baseline-implementations}, including information about the parameters for ATEMPO and the forecasting models that are not already stated in Section~\ref{sec:results}.

% \subsubsection*{Author Contributions}
% If you'd like to, you may include  a section for author contributions as is done
% in many journals. This is optional and at the discretion of the authors.

% \subsubsection*{Acknowledgments}
% Use unnumbered third level headings for the acknowledgments. All
% acknowledgments, including those to funding agencies, go at the end of the paper.

\bibliography{iclr2027_conference}
\bibliographystyle{iclr2027_conference}
\clearpage
\newpage
\appendix
\section{Dataset Details}
\label{sec:appendix-dataset-details}
\begin{table}[ht]
\setlength{\tabcolsep}{2pt}
\caption{Details of our suite of synthetic datasets with ground truth TE}
\label{tab:dataset}
\small
\begin{center}
\begin{tabular}{ccccccc}
\multicolumn{1}{c}{\bf Family} &\multicolumn{1}{c}{\bf Dataset}  &\multicolumn{1}{c}{\bf Source} &\multicolumn{1}{c}{\bf Number of} &\multicolumn{1}{c}{\bf Linear/} &\multicolumn{1}{c}{\bf Ground} &\multicolumn{1}{c}{\bf Known}\\
 &&\multicolumn{1}{c}{\bf Dimension} &\multicolumn{1}{c}{\bf true lags} &\multicolumn{1}{c}{\bf Non-linear} &\multicolumn{1}{c}{\bf Truth TE} &\multicolumn{1}{c}{\bf Lag(s)}
\\ \hline \\
AGM TE & flat\_linear & 1 & 1 & linear \\
\citep{pmlr-v275-kornai25a}& flat\_joint~& 1 & 1 & non-linear & $\checkmark$ & $\checkmark$ \\
\midrule
\multirow[l]{2}{*}{\makecell{multi-\\dimensional}} & lag\_linear\_m3\_d4& 5 & 1 & linear & $\checkmark$ & $\checkmark$ \\
& lag\_nonlin\_m3\_d4  &5 & 1 & non-linear & $\checkmark$ & $\checkmark$\\
& lag\_multisource & 5 & 1 & linear & $\checkmark$ & $\checkmark$ \\
\midrule
\multirow[l]{2}{*}{mean-blind} & lag\_sq\_m20 & 1 & 1 & non-linear & $\checkmark$ & $\checkmark$ \\
& lag\_vol\_m20 & 1 & 1 & non-linear & $\checkmark$ & $\checkmark$\\
\midrule
\multirow[l]{3}{*}{\makecell{time-varying\\lag}} &
switch & 1 & 2 (quick change) & linear & \(\times\) & $\checkmark$\\
& drift & 1 & 4 (gradual change) & linear & \(\times\) & $\checkmark$ \\
& null & 1 & 1 (no change) & linear & \(\times\) & $\checkmark$\\
\midrule
\multirow[l]{2}{*}{\makecell{time-varying\\mean-blind}} & switch\_sq & 1 & 2 (quick change) & non-linear & \(\times\) & $\checkmark$\\
& switch\_sq\_wk & 1 & 2 (quick change) & non-linear & \(\times\) & $\checkmark$\\
\bottomrule
\end{tabular}
\end{center}
\end{table}
\paragraph{Synthetic Datasets}
Table~\ref{tab:dataset} lists our suite of synthetic datasets and their properties. The datasets flat linear and flat joint are systems equivalent to~\citet{pmlr-v275-kornai25a}'s linear-VAR and joint process respectively, and we then constructed two multi-dimensional, two mean-blind, three with time-varying lag and two with both mean-blind and time-varying lag datasets.
The seven non-time-varying datasets are stationary, because an exact value of \(TE(k,l)\) is only tractable for stationary processes. The additional five datasets are used for lag detection and forecasting comparisons only. Datasets only have one informative dimension, but some have up to four redundant dimensions added. We vary the systems coupling nonlinearity, dimensionality and number of delays, and every system is roughly Gaussian and near-symmetric. We selected these families as they target the known limitations of existing TE estimators, high-dimensionality, non-linearity and lags that vary over time. All code to generate our datasets can be found in our GitHub\footnote{\code \label{fn:code}}.

\section{Experimental Protocol}
\subsection{Datasets}
\label{sec:appendix-dataset-implementations}
\paragraph{WEATHER5K}
We retain the original WEATHER-5K split using 2014-2021 for training, 2022 for validation and 2023 exclusively for testing. A fixed random subset of 300 stations is drawn once using the training data and used for all lag selection experiments. ATEMPO is trained separately for each of the \(5\times 5=25\) ordered source-target variable pairs, with maximum source-history depth \(K=336\)h and target conditioning depth \(L=24\)h, pooling windows across the 300 stations.
Each fitted model yields the full surface \(TE(k,l,t)\) but lag selection reads a single column of it: the marginal gain \(\delta_k(l)\) of Eq.~\ref{eq:profile} at the deepest conditioning depth, \(L=24\)h. The per-pair profiles are clipped at zero and normalized to unit sum, so that no high-TE pair dominates, then averaged over the 25 ordered pairs, and the (n) lags with the largest pooled gain form the selected set. For a variable against itself the gain at \(l = 24\) is zero for \(k \le 24\) by construction, so sub-daily self-lags enter the ranking through the cross-variable pairs only; the forecaster receives the target's recent history regardless.

Lag selection uses only these fitted models and training data; neither validation nor test observations are used to select lags. The selected source lags are then fixed and used as input features across stations. We evaluate the same selections by training forecasting models on both the 300-station subset and the complete 5,672-station dataset, allowing us to test whether lag structure estimated from the subset transfers across the full station population. Forecasting follows the benchmark parameters: iTransformer is trained with early stopping using the validation set, while the ridge penalty is selected on validation performance (ridge parameters are the same as specified in the Synthetic Datasets section below). All reported MSEs are evaluated on the held-out 2023 test set and averaged across the five meteorological variables. We use iTransformer because it performs competitively in the original benchmark while, unlike architectures requiring uniformly spaced consecutive inputs, it can directly consume the non-consecutive lag sets selected by ATEMPO.
%We use the $\boldsymbol{\delta}(l)=(\delta_1(l),\dots,\delta_K(l))$ derived from the \(TE(k,l,t)\) surface to rank the lags across five weather variables and compare this to the benchmark default input of 48 hours, corresponding to 240 input features.
% For each pair we take the marginal gain \(delta_{ij}(k)=\max{0,,\mathrm{TE}{ij}(k,L)-\mathrm{TE}{ij}(k{-}1,L)}\), average over 25 pairs, and keep the top-\(n\) lags, conditioning distable, so the score reflects the training years as a whole.
%WEATHER-5K compares mostly transformer-based architectures, as they are the most competitive compared to the ECMWF-HRES baseline and their PhysicsFormer model~\citep{weather5k}. 
%We choose the iTranformer architecture specifically because it consistently performs in the top three across variables and its architecture can handle being fed non-consecutive time-series input features. Pyraformer outperforms it in the original benchmark, however its pyramidal attention requires a continuous input, so it cannot represent a non-consecutive lag set. Similarly the PhysicsFormer model integrates over the uniformly-spaced hourly steps and the physics losses are defined over consecutive timesteps, so it also cannot handle irregular gaps. 

\paragraph{Neural Latent Benchmark}
The first two experiments use \textbf{MC\_RTT}: a 10.8-minute continuous recording of a monkey
performing random-target reaching, with finger position/velocity and 130 sorted units, binned at 20\,ms.
The benchmark labels every 600\,ms chunk of the public file as \emph{train} or \emph{val}; all fitting
and selection use the train chunks and all reported $R^2$ the val chunks. The third experiment uses \textbf{MC\_Maze}: 2{,}295 delayed maze-reaching trials with 182 neurons from M1/PMd, binned at 20\,ms, with hand velocity as the behavioral signal.

For NLB velocity, the aim is to forecast finger velocity $(v_x, v_y)$ 200\,ms ahead ($h = 10$ bins) from a base of the velocity's own past (10 bins) and the target-relative-to-finger position at lag~0, plus a budget of $B$ further lags of that input chosen by each selector from lags 1--29 (20--580\,ms). ATEMPO is fitted on the train rows (source = the 2-D relative-target signal, $K = 30$, conditioning on $l = 5$ bins of the velocity's past, 800 epochs), with a shifted-source null (source rolled by 3{,}000 bins, 60\,s); lags are ranked by null-corrected cumulative gains with lag~0 excluded, since it is already in the base. Every selection is scored by the same ridge ($\lambda = 1$) fitted on the train chunks; $R^2$ is the per-dimension coefficient of determination on the val chunks, averaged over $v_x$ and $v_y$.

In NLB decoding, the decoder is ridge ($\lambda = 10$) on all 130 smoothed units at the current bin (the base) plus $B$ selected lags from 1--19 (20--380\,ms of neural lead); no own-past input, as in the benchmark. Because the decoder has no access to the velocity's history, the matching ATEMPO quantity is checked in two forms: a fit conditioned on the velocity's past and a noise-conditioned fit whose estimand reduces to the unconditional information $I(v_t; X)$ the benchmark rewards (both: source = the first 3 population PCs, $K = 20$, 800 epochs, shifted-source nulls); the reported arm uses the conditioned fit's cumulative gains. 

Finally in MC\_Maze the task decodes hand velocity in the first 200\,ms after movement onset from the 20 leading population PCs at $B$ selected lag bins (20--400\,ms of neural lead), with no base input: at onset all information must come
from preparatory and early-motor activity at the right lead times, so the no-input decoder sits at $R^2 \approx 0$ and lag choice matters most here. The ATEMPO readout is one fit conditioned on trial phase (seven phases from pre-target to late movement, via the estimator's series-time conditioning; source = the first 3 PCs, $K = 20$, $l = 3$, 400 epochs, on a stratified subsample of 60k bins), with a shifted-source null (source rolled by 5{,}000 bins, 100\,s); the selection uses the movement-onset phase's null-corrected gains. Decoding uses a ridge ($\lambda = 1$) fitted and scored on disjoint trials
(80/20 split by trial id); $R^2$ is averaged over $v_x$ and $v_y$ on the held-out trials, within the onset phase.

\paragraph{Synthetic Datasets}
We split the data chronologically, with the first 70\% of points as train, 15\% for validation, to choose model parameters, and the final 15\% as test for measuring MSE. The forecaster model is a ridge regression with \(\alpha \in \{0.1,1,10,100 \}\) plus an MLP regressor with (32,) and (64,32) which uses early stopping. The input feature used for forecasting are the targets own past with a lag of l=3, \(y[t-l]\) for \(l\in{1,2,3}\), and the source \(x[t-k]\) where k is the detected lag at time t. Each (model,window size) pair is trained and averaged across three seeds, apart from ATEMPO where only one model is trained for all window sizes.

For evaluating performance at different window sizes in Figure~\ref{fig:ablation} and our ablation Table~\ref{tab:ablation}, we use three synthetic systems: switch, switch\_sq and switch\_sq\_wk. Each is a single recording of $N = 8{,}000$ samples generated with seed 0, in which the source-to-target delay is $m = 2$ for the first $4{,}000$ samples and $m = 5$ for the second. Each recording is cut into $W \in {8, 16, 32, 64}$ non-overlapping windows of $1{,}000$, $500$, $250$ and $125$ samples. 
For lag selection we pool the gain over conditioning depths,
\begin{equation}
  \bar\delta_k \;:=\; \frac{1}{L}\sum_{l=1}^{L}\delta_k(l), \qquad
  \hat k \;:=\; \arg\max_{k\in\{1,\dots,K\}} \bar\delta_k ,
  \label{eq:detect}
\end{equation}
so that a lag is credited only for information it adds at every depth of the target's own history; $\bar\delta_k \ge 0$ follows from Eq.~\ref{eq:profile}.
Every method reports one lag per window, the argmax of the pooled gain $\bar\delta_k$ of Eq.~\ref{eq:detect} with \(K=8\) and \(L=2\), and is scored against the planted lag at the window center. Detection accuracy is the fraction of correct windows, averaged over the three systems; error bars in Figure~\ref{fig:ablation} are the standard deviation across systems.

The three arms differ only in what data a fit sees.
\begin{itemize}
    \item Refit per window: the method is re-estimated on each window's samples alone, one estimate per $(k, l)$ cell, i.e.\ $K \cdot L \cdot W$ estimates per recording (a trained network per cell for TENDE, MINE, CCMI and NJEE; a plug-in estimate for NPEET, PCMI, Liang--Kleeman and DTW), or $W$ amortised fits for ATEMPO without the time axis. 
    \item Fit once: the method is estimated once on all $8{,}000$ samples and its single lag is applied to every window; without a time axis this is the only other route a baseline has, and it cannot exceed $0.5$ when the lag takes two values for equal durations.
    \item ATEMPO:  one amortized fit on all $8{,}000$ samples with the score network conditioned on series time, evaluated per window by restricting the expectation to that window's rows. All ATEMPO and TENDE fits use 400 epochs.
\end{itemize}

For forecasting results in Table~\ref{tab:ablation} and Table~\ref{tab:A_ablation_forecast}, we re-use the detected lags used to calculate lag detection accuracy in Table~\ref{tab:ablation} and Table~\ref{tab:A_window_length_detection}. 
For each system we draw three independent realizations of the same process (seeds 500--502), so no detector is scored on data it was fitted on; the lag schedule is identical, so a lag sequence detected on the seed-0 recording applies unchanged. A forecaster predicts $y_t$ from its own three most recent values plus the single source value $x_{t - \hat k(t)}$, where $\hat k(t)$ is the lag the method detected for the window containing $t$. Within each $4{,}000$-sample half, the first 70\% of rows are used for fitting, the next 15\% to choose between ridge regression (five penalties, $10^{-2}$ to $10^{2}$) and a small MLP (one or two hidden layers), and the last 15\% for the reported test MSE. The oracle is the identical pipeline given the true lag in every row; we report the ratio of test MSE to oracle test MSE, averaged over seeds and then over systems. 

\paragraph{MMTT-Bench}
We concatenate the benchmark's train, validation and test splits into one series of 2048 steps. The target takes four discrete values \(\{-1, -0.5, 0, 1 \}\); since density-based estimators require a continuous target, we add Gaussian dither with \(\sigma = 0.02\). The benchmark's text at step \(t\) describes step \(t+1\) (a native lag of 1). We plant a delay \(m\) that advances the text stream by \(m-1\) steps. Four conditions are run: the benchmark's correct, incorrect and irrelevant text, plus a shuffled null in which the correct text is permuted in time on the source side only, leaving the target's own dynamics intact. Sentences are embedded with the benchmark's default encoder, sentence-transformers/all-distilroberta-v1, reduced by PCA and standardized. Detection and forecasting use different PCA dimensions because their sample budgets differ: \(d=4\) for TE estimation (\(K=8\) lags \(\times 4\) dims \(=32\) source features against around 2000 samples) and \(d=16\) for the ridge forecaster (128 features against around 1400 training rows). 

ATEMPO is fitted once per (condition, planted lag, seed) with \(K=8\) source lags and \(L=3\) target lags, 400 epochs, and otherwise the defaults of the ridge model used in Synthetic Datasets above. The downstream forecasting model predicts \(y_t\) from its own three most recent values plus the 16-dimension text embedding at a window of \(W\) consecutive lags. The first 70\% of the data is used for fitting, with the final 20\% of this for validation to select the ridge penalty. The last 30\% of the data is the held out test set for calculating forecasting MSE.

\subsection{Baselines}
\label{sec:appendix-baseline-implementations}
We implement eight TE estimator baselines to compare to ATEMPO. Details of our implementations are outlined below, with full code available in our GitHub\footref{fn:code}. As Liang-Kleeman and DTW do not estimate transfer entropy in nats directly, we do not evaluate them for accuracy against ground truth TE. Instead we report their output matrix as a cumulative sum over source lags, such that when the \(\Delta\) is taken it corresponds to the per-lag quantity, equivalent to the \(\Delta TE\) surface itself.

\begin{itemize}[leftmargin=*]
    \item \textbf{NPEET}: Non-parametric k-nearest-neighbor CMI computes CMI from k-NN distance statistics in the join space. We use the NPEET python mutual information estimator implementation based on~\citet{kraskov_estimating_2004} with \(k=3\) neighbors and convert its base-2 output to nats. It needs no training, making it the cheapest baseline, but it is also non-deterministic. Given no seed, repeated calls on identical data can differ in the third decimal.
    \item \textbf{NJEE}: Neural joint entropy estimation estimates transfer entropy as a difference of two conditional entropies \(TE = H(Y \mid Z) - H(Y \mid X,Z)\), each obtained as a neural classifier's cross-entropy, which upper-bounds the conditional entropy and tightens as the classifier improves. We discretize the target into 8 equally-probable levels so each entropy becomes a classification loss, leaving the conditional variables continuous, and train a two-hidden-layer MLP (128 units, Adam 400 steps) per term. Cross-entropy is evaluated on a held-out split, because in-sample cross-entropy underestimates entropy and would inflate every TE, and because TE is a difference of two terms the leading discretization bias cancels. For multivariate targets, we only discretize the first target dimension, making its value a lower bound on the true TE.
    \item \textbf{MINE}: The Donsker-Varadhan neural mutual information estimator~\citep{belghazi2018mutual} where a critic \(T_\theta\) is trained to maximize \(I(A;B) \geq \mathbb{E}{p(a,b)}[T] - \log \mathbb{E}{p(a)p(b)}[e^{T}]\). We use a two-hidden-layer ReLU MLP (128 units) trained with Adam (learning rate \(10^{-3}\), 400 steps, batch 256), drawing marginal samples by shuffling (B) within each batch, and evaluate the bound on the full sample at the full sample at the final step. Because MINE estimates an unconditional MI, transfer entropy is obtained as a difference of two separately estimated quantities: \((TE = I(Y; X, Z) - I(Y; Z))\). The errors therefore add rather than cancel, which is the decomposition that the next baseline, CCMI, was introduced to avoid.
    \item \textbf{CCMI}: The classifier-based conditional mutual information estimator (CCMI)~\citep{mukherjee2020ccmi} estimates \(I(Y; X \mid Z)\) directly. A binary classifier is trained to separate the joint \(p(x,y,z)\) from the conditional, producing \(p(x \mid z)p(y \mid z)p(z)\), and the Donsker-Varadhan functional applied to its logits, which are pointwise log-density ratio estimates, gives the CMI without the difference of large terms. We construct the conditional-product sample by the local permutation scheme of~\citep{runge2018conditional}: for each \(i\), \(x_i\) is replaced by \(x_j\) drawn from among \(i\)'s five nearest neighbors in \(Z\). A global shuffle would destroy the \(X-Z\) relationship and inflate the estimate. Classifier and optimizer settings are the same as MINE.
    \item \textbf{Liang-Kleeman}: Liang-Kleeman information flow is a causality measure derived from the rate of change of the target's marginal entropy once its own self-influence is removed~\citep{liang2013liang}. We implement the maximum-likelihood estimator of~\citet{liang2014unraveling} in its multivariate form: for the linear stochastic system \(\dot{x}i = \sum_j a{ij} x_j + \varepsilon\), the coefficients are \(\hat{a} = C^{-1} C_d\) with \(C\) the state covariance and \(C_{d}[k,i] = \text{Cov}(x_k, \dot{x}i)\) using an Euler forward difference, and the flow is \(T{j \to i} = a_{ij} C_{ij} / C_{ii}\). At two components this reduces algebraically to the published bivariate closed form \(T_{2\to1} = (C_{11}C_{12}C_{2,d1} - C_{12}^2 C_{1,d1}) / (C_{11}^2 C_{22} - C_{11}C_{12}^2)\), which we verify numerically. Because the forward difference makes a source sample at \(t\) act on the target at \(t+1\), probing delay \(k\) shifts the source by \(k-1\). The final result is a rate in nats per unit time, which is not an estimator of transfer entropy directly,  , so it is only judged on lag discovery and downstream forecasting. It also has no target-history-depth parameter, so its matrix is constant along \(l\).
    \item \textbf{DTW}: Dynamic time warping is a lead-lag heuristic rather than an information measure~\citep{dtw-berndt-clifford}. We align the two series with a Sakoe--Chiba band of half-width 64~\citep{sakoe1978} and read the offset \(j-i\) of every matched pair as a local estimate of how far the target trails the source. The histogram of these offsets over lags \(1..K\), as a fraction of matched pairs, is the per-lag profile, and its mode is the detected delay. Both series are z-scored and first-differenced before alignment, following the derivative variant of~\citet{keogh2001}: raw DTW is biased by exactly one lag on autoregressive targets, because the target's own level memory drags the alignment forward a step, while differencing leaves the innovation the coupling acts on (top-1 accuracy on the synthetic suite rises from 0.66 to 0.98). The band is fixed rather than tied to \(K\) so that every cell of the matrix reads the same alignment and only the histogram window moves. The output is dimensionless and not a bound on TE, so DTW is excluded from the TE error comparison; and because DTW matches shapes without conditioning on the target's own past, it has no \(l\) axis and cannot separate directed transfer from a shared driver, so its lag should be read as ``these shapes line up at this offset'', not as evidence of a directed link.
    \item \textbf{PCMI}: Permutation conditional mutual information (PCMI)~\citep{li2011characterization} computes  \(I(Y_t; X_{t-k} \mid Y_{t-1..t-l})\) on ordinal patterns rather than the values themselves. Each series is symbolized by the Bandt-Pompe encoding~\citep{bandt2002permutation} at embedding dimension \(m=3\) and delay \(\tau=1\) using windows ending at \(t\) so that a symbol depends only on the series up to \(t\). The conditional mutual information is formed from the entropies: \(I(Y;X\mid Z) = H(Y,Z) + H(X,Z) - H(X,Y,Z) - H(Z)\), each with the Miller-Madow bias correction \((\hat{K}-1)/2N\)~\citep{paninski2003estimation}. It estimates the CMI of the symbol sequences, with upper bound \(\log m! = 1.79\) nats at \(m=3\). Its lag resolution is also limited by \(m\), where ordinal windows at delays within \(m\) of another share source samples, so a delay of \(k=3\) leaves PCMI non-zero at lags 1 and 2. 
\end{itemize}

\section{Attribution of Transfer Entropy Across Delays}
\label{app:profile}

Equation~\ref{eq:profile} assigns to each source delay the increase in transfer entropy obtained by admitting it after all shorter delays. This appendix locates that definition among its alternatives. We show that the two attributions in common use are not decompositions of $\mathrm{TE}^{X\to Y}(K,l)$, that the increment and the Shapley value both are, and that these two differ in their treatment of redundant delays in a manner that determines which is appropriate for delay estimation.

Let $N:=\{1,\dots,K\}$ index the source delays, $\mathbf{X}_S:=(X_{t-j})_{j\in S}$ for $S\subseteq N$, and define the set function
\begin{equation}
  v(S)\;:=\;I\bigl(Y_t;\mathbf{X}_S\mid\mathbf{Y}^{(l)}_t\bigr),
  \qquad
  v(\emptyset)=0,
  \qquad
  v(N)=\mathrm{TE}^{X\to Y}(K,l).
  \label{eq:app-game}
\end{equation}
The chain rule gives $v(S\cup\{j\})-v(S)=I(Y_t;X_{t-j}\mid\mathbf{X}_S,\mathbf{Y}^{(l)}_t)\ge0$, so $v$ is monotone. An attribution assigns a share $\varphi_k$ of $v(N)$ to each delay, and is a \emph{decomposition} when $\sum_{k\in N}\varphi_k=v(N)$ with $\varphi_k\ge0$.

For a permutation $\pi$ of $N$, let $P^\pi_k$ denote the delays preceding $k$ in $\pi$ and define the marginal vector
\begin{equation}
  m^{\pi}_k\;:=\;v\bigl(P^\pi_k\cup\{k\}\bigr)-v\bigl(P^\pi_k\bigr),
  \qquad\text{so that}\qquad
  \sum_{k\in N} m^{\pi}_k\;=\;v(N)
  \label{eq:app-marginal}
\end{equation}
by telescoping, for every $\pi$. Monotonicity makes each term non-negative, so every marginal vector is a decomposition. This is the property the two common attributions lack. Scoring delays in isolation and conditioning on all remaining delays,
\begin{equation}
  m_k(l)\;=\;v\bigl(\{k\}\bigr)-v(\emptyset),
  \qquad
  c_k(l)\;=\;v(N)-v\bigl(N\setminus\{k\}\bigr),
  \label{eq:app-alternatives}
\end{equation}
are the marginal contributions of $k$ when $k$ is placed first and last respectively. Neither corresponds to a single permutation, since every delay cannot be ordered first, nor every delay last, and Eq.~\ref{eq:app-marginal} therefore does not apply to either.

The consequence is visible whenever delays carry overlapping information, which an autocorrelated source produces at every pair of adjacent delays. Let $X_{t-1}=X_{t-2}=Z$ with $I(Y_t;Z\mid\mathbf{Y}^{(l)}_t)=1$ nat, so $v(\{1\})=v(\{2\})=v(\{1,2\})=1$. Then $(m_1,m_2)=(1,1)$, which sums to twice $v(N)$ and credits the shared nat to both delays; in a recording this appears as a peak broadened across all delays adjacent to the true one. Conversely $(c_1,c_2)=(0,0)$, which sums to zero, because each copy of $Z$ renders the other redundant and no delay adds anything to the remainder of the history.

Among decompositions the choice is between a single ordering and an average over orderings. The increment of Eq.~\ref{eq:profile} is the marginal vector at the recency ordering $\pi_0=(1,2,\dots,K)$, while the Shapley value is the mean over all orderings,
\begin{equation}
  \delta_k(l)\;=\;m^{\pi_0}_k,
  \qquad
  \phi_k(l)\;=\;\frac{1}{K!}\sum_{\pi} m^{\pi}_k
  \;=\!\!\sum_{S\subseteq N\setminus\{k\}}\!\!\frac{|S|!\,(K-|S|-1)!}{K!}\Bigl[v\bigl(S\cup\{k\}\bigr)-v(S)\Bigr].
  \label{eq:app-shapley}
\end{equation}
On the system above these give $(\delta_1,\delta_2)=(1,0)$ and $(\phi_1,\phi_2)=(0.5,0.5)$. Both are decompositions, and they differ only in how the redundant nat is divided. Averaging over orderings makes $\phi$ invariant to the labelling of delays, so a delay and a later repetition of it receive equal shares. Delays are not exchangeable: influence propagates forward, and information present at delay $1$ and repeated at delay $2$ originates at delay $1$. Fixing $\pi_0$ imposes precisely this precedence, which is what permits $\boldsymbol{\delta}(l)$ to be read as a statement about when the source acts. Evaluating Eq.~\ref{eq:app-shapley} would in any case require $2^{K-1}$ coalitions per delay, and at the depths used here ($K=336$ on WEATHER-5K) no sampling scheme attains the precision a delay estimate requires.

%A final consequence concerns estimation rather than definition. Since $\delta_k(l)$ differences two cells of the same surface, an estimator fitted separately per cell contributes two statistically independent errors to every entry of the profile, whereas ATEMPO evaluates both cells with one network, where the errors are correlated and cancel in part. Amortisation therefore affects the accuracy of delay detection and not only its cost.
A final consequence concerns estimation rather than definition. Since $\delta_k(l)$ differences two cells of the same surface, an estimator fitted separately per cell contributes two independent errors to every entry of the profile, whereas a single amortized network evaluates both cells with shared parameters and the errors are correlated. Table~\ref{tab:A_profile_error_correlation} measures this on the stationary benchmark, where both estimators were run on the same systems and seeds against exact ground truth. Across five seeds, the errors of adjacent cells correlate at $0.05$ for per-cell fits and $0.72$ for ATEMPO; the variance of the error in $\delta_k$ is $0.95$ of the sum of the two cell variances for per-cell fits, as independence predicts, and $0.29$ for ATEMPO. As a result the profile is noisier than the surface for per-cell fits (sd $0.026$ against $0.019$ nats) and less noisy than the surface for ATEMPO ($0.008$ against $0.011$), a three-fold reduction in the sd of $\delta_k$. Amortization therefore affects the accuracy of delay detection and not only its cost.
\begin{table}[t]
\centering
\small
\caption{Do errors in adjacent cells of the TE surface cancel when differenced? Stationary benchmark at $K=6$, $L=3$, same (system, seed) pairs for both estimators, exact ground truth. Across seeds, per cell: the sd of the TE estimate, the sd of the delay profile $\delta_k = \mathrm{TE}(k,l)-\mathrm{TE}(k-1,l)$, the ratio of the variance of the $\delta_k$ error to the sum of the two cell variances (1 for independent errors, below 1 when errors are positively correlated), and the pooled correlation of the two cells' errors. The null direction has true TE $=0$ everywhere, so its errors are pure estimation noise.}
\label{tab:A_profile_error_correlation}
\begin{tabular}{llccccccc}
\toprule
 &  & systems & seeds & cells & sd & sd & Var($\delta$ err) / & corr(err$_k$, \\
direction & estimator &  &  &  & TE(k)  & $\delta_k$  & (Var$_k$ + Var$_{k-1}$) & err$_{k-1}$) \\
\midrule
\multirow[t]{2}{*}{causal} & \makecell[l]{ATEMPO\\(one fit)} & 5 & 5 & 75 & 0.011 & 0.008 & 0.286 & 0.715 \\
 & \makecell[l]{TENDE\\(one fit per cell)} & 5 & 5 & 75 & 0.019 & 0.026 & 0.947 & 0.053 \\
\cline{1-9}
\multirow[t]{2}{*}{null} & \makecell[l]{ATEMPO\\(one fit)} & 5 & 5 & 75 & 0.001 & 0.000 & 0.222 & 0.782 \\
 & \makecell[l]{TENDE\\(one fit per cell)} & 5 & 5 & 75 & 0.001 & 0.001 & 0.715 & 0.300 \\
\cline{1-9}
\bottomrule
\end{tabular}
\end{table}

\clearpage
\newpage
\section{Synthetic Dataset Results}
\label{sec:appendix-synthetic-results}
\begin{table}[ht]
\centering
\tiny
\caption{Relative error (\%) of the TE estimate at the true lag, mean $\pm$ sd over 5 seeds. Bold is the lowest error per system.}
\label{tab:A_te_relerr_by_system}
\begin{tabular}{lllllllll}
\toprule
method & family & NPEET & NJEE & MINE & CCMI & PCMI & TENDE & ATEMPO \\
system &  &  &  &  &  &  &  &  \\
\midrule
flat\_linear & AGM-TE & 6.7 ± 2.3 & 18.5 ± 4.5 & 8.5 ± 1.4 & 5.1 ± 5.2 & 67.6 ± 2.6 & \textbf{3.9} ± 4.7 & 6.8 ± 4.1 \\
flat\_joint & AGM-TE & \textbf{2.1} ± 1.7 & 19.0 ± 3.7 & 12.1 ± 7.1 & 2.8 ± 2.5 & 76.0 ± 1.4 & 8.7 ± 3.0 & 9.7 ± 5.2 \\
lag\_linear\_m3\_d4 & multi-dimensional & 93.1 ± 5.3 & 38.6 ± 4.7 & 8.8 ± 4.2 & 79.4 ± 9.4 & 27.0 ± 6.3 & 6.1 ± 2.8 & \textbf{3.7} ± 2.6 \\
lag\_multisource & multi-dimensional & 44.9 ± 2.7 & 30.3 ± 5.6 & 7.4 ± 4.4 & 20.6 ± 4.5 & 27.2 ± 4.3 & \textbf{2.5} ± 1.6 & 4.9 ± 0.9 \\
lag\_nonlin\_m3\_d4 & multi-dimensional & 81.8 ± 2.1 & 35.2 ± 3.3 & 18.8 ± 3.7 & \textbf{8.7} ± 4.0 & 47.6 ± 2.2 & 29.6 ± 30.9 & 9.6 ± 3.0 \\
lag\_sq\_m20 & mean-blind & 90.8 ± 0.4 & 31.8 ± 2.2 & 26.5 ± 2.7 & 20.9 ± 4.1 & 81.6 ± 0.8 & 28.7 ± 3.0 & \textbf{13.4} ± 3.0 \\
lag\_vol\_m20 & mean-blind & 72.3 ± 3.4 & 57.1 ± 7.8 & \textbf{1.2} ± 1.4 & 30.6 ± 5.8 & 59.1 ± 3.3 & 34.6 ± 9.0 & 9.9 ± 3.3 \\
\bottomrule
\end{tabular}
\end{table}

\begin{table}[ht]
\centering
\tiny
\caption{Signed relative error (\%) at the true lag (negative = under-estimate), mean $\pm$ sd over seeds.}
\label{tab:A_te_signed_err_by_system}
\begin{tabular}{llllllllll}
\toprule
method & family & true TE & NPEET & NJEE & MINE & CCMI & PCMI & TENDE & ATEMPO \\
system &  &  &  &  &  &  &  &  &  \\
\midrule
flat\_linear & AGM-TE & 0.128 & -0.6 ± 7.8 & -18.5 ± 4.5 & 2.1 ± 9.3 & -1.9 ± 7.5 & -67.6 ± 2.6 & -3.3 ± 5.3 & 2.3 ± 8.3 \\
flat\_joint & AGM-TE & 0.415 & -0.7 ± 2.8 & -19.0 ± 3.7 & -12.1 ± 7.1 & -2.0 ± 3.3 & -76.0 ± 1.4 & -8.7 ± 3.0 & -9.1 ± 6.3 \\
lag\_linear\_m3\_d4 & multi-dim. & 0.128 & -93.1 ± 5.3 & -38.6 ± 4.7 & 8.8 ± 4.2 & 79.4 ± 9.4 & -27.0 ± 6.3 & -6.1 ± 2.8 & 1.5 ± 4.6 \\
lag\_multisource & multi-dim. & 0.128 & -44.9 ± 2.7 & -30.3 ± 5.6 & -6.3 ± 6.2 & 20.6 ± 4.5 & -27.2 ± 4.3 & 0.5 ± 3.2 & 2.9 ± 4.6 \\
lag\_nonlin\_m3\_d4 & multi-dim. & 0.415 & -81.8 ± 2.1 & -35.2 ± 3.3 & -18.8 ± 3.7 & 8.7 ± 4.0 & -47.6 ± 2.2 & -29.6 ± 30.9 & -9.6 ± 3.0 \\
lag\_sq\_m20 & mean-blind & 0.988 & -90.8 ± 0.4 & -31.8 ± 2.2 & -26.5 ± 2.7 & -20.9 ± 4.1 & -81.6 ± 0.8 & -28.7 ± 3.0 & -13.4 ± 3.0 \\
lag\_vol\_m20 & mean-blind & 0.170 & -72.3 ± 3.4 & -57.1 ± 7.8 & -0.4 ± 1.9 & 30.6 ± 5.8 & -59.1 ± 3.3 & -34.6 ± 9.0 & -9.9 ± 3.3 \\
\bottomrule
\end{tabular}
\end{table}
\begin{table}[ht]
\centering
\tiny
\caption{Top-1 lag recovery (successes / trials) on every synthetic system. Stationary systems: one trial per seed; time-varying systems: one trial per 1000-sample window.}
\label{tab:A_lag_detection_by_system}
\begin{tabular}{lllccccccccc}
\toprule
 family & system & trials & NPEET & NJEE & MINE & CCMI & PCMI & Liang & DTW & TENDE & ATEMPO \\
&  &  &  &  &  &  &  & -Kleeman &  &  &  \\
\midrule
 \multirow[t]{2}{*}{AGM-TE} & flat\_linear &  seeds & 5/5 & 5/5 & 5/5 & 5/5 & 5/5 & 5/5 & 5/5 & 5/5 & 5/5 \\
& flat\_joint & seeds & 5/5 & 5/5 & 5/5 & 5/5 & 5/5 & 2/5 & 5/5 & 5/5 & 5/5 \\
\multirow[t]{3}{*}{\makecell[c]{multi-\\dimensional}} &
lag\_linear\_m3d4 & seeds & 1/5 & 5/5 & 5/5 & 5/5 & 5/5 & 5/5 & 5/5 & 5/5 & 5/5 \\
& lag\_multisource & seeds & 5/5 & 5/5 & 5/5 & 5/5 & 5/5 & 3/5 & 5/5 & 5/5 & 5/5 \\
& lag\_nonlin\_m3d4  & seeds & 5/5 & 5/5 & 5/5 & 5/5 & 5/5 & 3/5 & 5/5 & 5/5 & 5/5 \\
\multirow[t]{2}{*}{mean-blind}
& lag\_sq\_m20 & seeds & 5/5 & 5/5 & 5/5 & 5/5 & 5/5 & 0/5 & 0/5 & 5/5 & 5/5 \\
& lag\_vol\_m20  & seeds & 5/5 & 5/5 & 5/5 & 5/5 & 5/5 & 0/5 & 0/5 & 5/5 & 5/5 \\
\multirow[t]{3}{*}{time-varying}
& tv\_switch  & windows & 8/8 & 7/8 & 8/8 & 7/8 & 7/8 & 3/8 & 5/8 & 8/8 & 8/8 \\
& tv\_drift & windows & 8/8 & 5/8 & 7/8 & 7/8 & 6/8 & 5/8 & 7/8 & 7/8 & 6/8 \\
& tv\_null  & windows & 8/8 & 6/8 & 8/8 & 6/8 & 8/8 & 4/8 & 4/8 & 8/8 & 8/8 \\
\multirow[t]{2}{*}{\makecell[c]{mean-blind\\time-varying}}
& tv\_switch\_sq  & windows & 8/8 & 8/8 & 8/8 & 8/8 & 8/8 & 1/8 & 8/8 & 8/8 & 8/8 \\
& tv\_switch\_sq\_wk  & windows & 6/8 & 6/8 & 6/8 & 7/8 & 3/8 & 2/8 & 4/8 & 5/8 & 8/8 \\
\bottomrule
\end{tabular}
\end{table}
\paragraph{TE Estimation}
Table~\ref{tab:A_te_relerr_by_system} shows the TE relative error across all seven datasets with known TE, which are averaged to produce the results in Table~\ref{tab:te-estimate}.
Table~\ref{tab:A_te_signed_err_by_system} also includes the sign of the TE estimation error, showing how some estimators like NPEET and NJEE consistently underestimate TE, whereas CCMI tends to overestimate, especially on multi-dimensional systems. 

\paragraph{Lag Detection Accuracy}
% \begin{figure}[h]
% \begin{center}
% \includegraphics[width=\textwidth]{figures/small_grid.png}
% \end{center}
% \caption{Lag detection accuracy over 5 or 8 seeds, across all baselines and synthetic datasets\label{fig:lag_estimation_grid}}
% \end{figure}

Table~\ref{tab:A_lag_detection_by_system} shows the Top-1-accuracy lag detection results broken down per dataset, where a success is only if the true lag has the highest \(\Delta TE\) estimate. Non-time-varying lags were evaluated across five seeds, and time-varying lags were evaluated with one seed and eight windows, so a score of 6/8 means the maximum TE corresponded to the true lag in six of the eight windows.

\subsection{Ablation}
\label{sec:appendix-ablation}
\begin{table}[ht]
\centering
\tiny
\caption{Top-1 detection rate on a moving lag against samples per window, per system and averaged. Baselines are re-fitted in every window; ATEMPO is fitted once.}
\label{tab:A_window_length_detection}
\begin{tabular}{l|rrrr|rrrr|rrrr}
\toprule
 & \multicolumn{4}{c}{switch} & \multicolumn{4}{|c|}{switch\_sq} & \multicolumn{4}{c}{switch\_sq\_wk} \\
win\_len & 125 & 250 & 500 & 1000 & 125 & 250 & 500 & 1000 & 125 & 250 & 500 & 1000 \\
\midrule
\makecell[l]{Liang-\\Kleeman} & 0.39 & 0.38 & 0.44 & 0.38 & 0.17 & 0.19 & 0.12 & 0.12 & 0.14 & 0.03 & 0.12 & 0.25 \\
NPEET & 0.67 & 0.84 & 0.94 & 1.00 & 0.95 & 1.00 & 1.00 & 1.00 & 0.42 & 0.69 & 0.81 & 0.75 \\
NJEE & 0.19 & 0.25 & 0.69 & 0.88 & 0.69 & 0.88 & 1.00 & 1.00 & 0.17 & 0.19 & 0.19 & 0.75  \\
MINE & 0.22 & 0.41 & 0.69 & 1.00 & 0.61 & 0.94 & 0.94 & 1.00 & 0.12 & 0.22 & 0.38 & 0.75 \\
CCMI & 0.03 & 0.03 & 0.06 & 0.88 & 0.44& 0.88 & 1.00 & 1.00 & 0.09 & 0.34 & 0.56 & 0.88 \\
PCMI & 0.33 & 0.44 & 0.69 & 0.88 & 0.39 & 0.81 & 1.00 & 1.00 & 0.16 & 0.19 & 0.31 & 0.38 \\
DTW & 0.48 & 0.53 & 0.75 & 0.62 & 0.62 & 0.72 & 1.00 & 1.00 & 0.19 & 0.19 & 0.44 & 0.50\\
TENDE & 0.06 & 0.06 & 0.25 & 1.00 & 0.14 & 0.16 & 0.75 & 1.00 & 0.12 & 0.12 & 0.06 & 0.62  \\
\makecell[l]{ATEMPO\\(no time-conditioning,\\per-window)} & 0.09 & 0.03 & 0.62 & 1.00 & 0.09 & 0.25 & 0.94 & 1.00 & 0.00 & 0.09 & 0.31 & 1.00  \\
\makecell[l]{ATEMPO\\(no time-conditioning)} & 0.59 & 0.56 & 0.50 & 0.50 & 0.55 & 0.50 & 0.50 & 0.50 & 0.78 & 0.75 & 0.81 & 0.75  \\
\bf{ATEMPO} & 1.00 & 1.00 & 1.00 & 1.00 & 1.00 & 1.00 & 1.00 & 1.00 & 1.00 & 1.00 & 1.00 & 1.00\\
\bottomrule
\end{tabular}
\end{table}
\begin{table}[ht]
\centering
\tiny
\caption{Ablation arms: forecasting MSE against window length, per system.}
\label{tab:A_ablation_forecast}
\begin{tabular}{l|rrrr|rrrr|rrrr}
\toprule
 & \multicolumn{4}{c}{switch} & \multicolumn{4}{|c|}{switch\_sq} & \multicolumn{4}{c}{switch\_sq\_wk} \\
win\_len& 125 & 250 & 500 & 1000 & 125 & 250 & 500 & 1000 & 125 & 250 & 500 & 1000 \\
\midrule
NPEET & 0.224 & 0.228 & 0.208 & 0.209 & 0.313 & \textbf{0.099} & \textbf{0.099} & \textbf{0.099} & 0.117 & 0.111 & 0.108 & 0.097 \\
NJEE & 0.257 & 0.256 & 0.249 & 0.210 & 0.578 & 0.292 & \textbf{0.099} & \textbf{0.099} & 0.124 & 0.122 & 0.122 & 0.117 \\
MINE & 0.251 & 0.244 & 0.248 & 0.209 & 0.899 & 0.132 & 0.114 & \textbf{0.099} & 0.123 & 0.126 & 0.111 & 0.098 \\
CCMI & 0.258 & 0.259 & 0.258 & \textbf{0.208} & 1.142 & 0.372 & \textbf{0.099} & \textbf{0.099} & 0.122 & 0.121 & 0.117 & 0.093 \\
PCMI & 0.257 & 0.260 & 0.279 & \textbf{0.208} & 1.226 & 0.402 & \textbf{0.099} & \textbf{0.099} & 0.122 & 0.124 & 0.122 & 0.116 \\
Liang-Kleeman & 0.246 & 0.244 & 0.236 & 0.257 & 1.446 & 1.442 & 1.433 & 1.446 & 0.122 & 0.122 & 0.122 & 0.119 \\
DTW & 0.258 & 0.245 & 0.234 & 0.262 & 0.607 & 0.748 & \textbf{0.099} & \textbf{0.099} & 0.122 & 0.122 & 0.114 & 0.106 \\
TENDE & 0.258 & 0.259 & 0.258 & 0.209 & 1.451 & 1.500 & 0.372 & \bf{0.099} & 0.122 & 0.122 & 0.122 & 0.114 \\
\makecell[l]{ATEMPO\\(no time-\\conditioning,\\per-window)} & 0.255 & 0.257 & 0.236 & 0.209 & 1.451 & 1.449 & 0.336 & \bf{0.099} & 0.122 & 0.123 & 0.114 & \bf{0.093} \\
\makecell[l]{ATEMPO\\(no time-\\conditioning)} & 0.245 & 0.263 & 0.272 & 0.247 & 0.954 & 0.955 & 0.955 & 0.955 & 0.103 & 0.107 & 0.100 & 0.115 \\
ATEMPO & \bf{0.209} & \bf{0.209} & \bf{0.209} & 0.209 & \bf{0.099} & \bf{0.099} & \bf{0.099} & \bf{0.099} & \bf{0.093} & \bf{0.093} & \bf{0.093} & \bf{0.093} \\
\bottomrule
\end{tabular}
\end{table}

Table~\ref{tab:A_window_length_detection} shows the full results across the three synthetic time-varying systems which were averaged to produce Table~\ref{tab:ablation} and Figure~\ref{fig:ablation}. ATEMPO has perfect detection across all datasets and all windows. ATEMPO (no time-conditioning) is robust to window size, but performance is significantly degraded without time-conditioning. ATEMPO (per-window) shows how having to train data per-window degrades performance significantly as window size decreases. This trend is also observed across all other baselines as they too have to be trained per-window. 

Table~\ref{tab:A_ablation_forecast} extends this to compare forecasting MSE across these datasets, with the full ATEMPO selected-lags always producing the lowest MSE which is robust to window size, whereas the ablations trained per-window are only able to match this performance at the largest window size. ATEMPO (no time-conditioning) represents ATEMPO with lag amortization but time-conditioning removed and always performs worst, showing the benefit time conditioning has to downstream forecasting.
\clearpage
\newpage
\section{MMTT-Bench Full Results}
\label{sec:appendix-mmtt-results}
\begin{table}[ht]
\centering
\caption{MMTT-Bench lag recovery per planted delay (3 seeds), null rejection on shuffled text, and cost relative to ATEMPO.}
\label{tab:A_mmtt_detection_all}
\begin{tabular}{lrrrrrrrrr}
\toprule
 & m=1 & m=2 & m=3 & m=4 & m=5 & m=6 & mean & shuffled & time rel. \\
method &  &  &  &  &  &  &  & (null) & ATEMPO \\
\midrule
\makecell[l]{Liang-\\Kleeman} & 0.000 & 0.000 & 0.000 & 0.000 & 0.000 & 0.000 & 0.000 & 0.940 & 0.000 \\
DTW & 0.000 & 0.000 & 0.000 & 0.000 & 0.000 & 0.000 & 0.000 & 0.890 & 0.000 \\
NPEET & 1.000 & 0.000 & 0.000 & 1.000 & 0.000 & 0.000 & 0.330 & 0.890 & 0.050 \\
NJEE & 1.000 & 0.330 & 1.000 & 1.000 & 0.330 & 0.670 & 0.720 & 0.940 & 0.550 \\
MINE & 1.000 & 0.330 & 1.000 & 1.000 & 1.000 & 0.670 & 0.830 & 0.940 & 1.330 \\
CCMI & 1.000 & 0.000 & 0.670 & 1.000 & 1.000 & 1.000 & 0.780 & 0.830 & 0.410 \\
PCMI & 1.000 & 1.000 & 1.000 & 1.000 & 1.000 & 1.000 & 1.000 & 0.890 & 0.000 \\
TENDE & 1.000 & 1.000 & 1.000 & 1.000 & 0.670 & 1.000 & 0.940 & 1.000 & 16.280 \\
ATEMPO & 1.000 & 0.670 & 1.000 & 1.000 & 1.000 & 1.000 & 0.940 & 1.000 & 1.000 \\
\bottomrule
\end{tabular}
\end{table}
\begin{table}[ht]
\caption{MMTT-Bench by text condition, averaged over planted delays $m\in\{1-6\}$ and 3 seeds. Left: what ATEMPO estimates. Right: \% MSE reduction over forecasting the signal from its own past alone, when one text lag is added: the benchmark's default lag 1, the lag ATEMPO selects, and a random lag. Only correct text carries a lag to find; on the three controls no choice of lag helps.}
\label{tab:T_mmtt_value}
\begin{tabular}{lccc|ccc}
\toprule
 &  \multicolumn{3}{c}{ATEMPO estimates} &  \multicolumn{3}{c}{\% MSE}\\
condition& TE (nats) & peak $\delta$TE & lag found & default (lag 1) & \textsc{atempo} lag & random lag \\
\midrule
correct & 0.52 & 0.27 & 1.00 & 15.59 & 41.26  & 10.37  \\
incorrect & 0.21 & 0.05 & 0.00 & 7.89 & 1.09  & 3.12  \\
irrelevant & 0.14 & 0.05 & 0.00 & -0.69 & -2.35  & -1.32  \\
shuffled & 0.15 & 0.05 & 0.00 & -1.50 & -1.18  & -1.19\\
\bottomrule
\end{tabular}
\end{table}
Table~\ref{tab:A_mmtt_detection_all} shows the full lag detection results for each planted lag and the shuffled condition where no lags are present. All these detection results are averaged to produce the column in result Table~\ref{tab:te-estimate}. ATEMPO and TENDE each only fail to detect the one shifted lag, narrowly behind PCMI which detects them all perfectly. 

Table~\ref{tab:T_mmtt_value} shows the peak \(\delta TE\) values for each of the different text categories in MMTT-Bench. ATEMPO correctly reconstructs the datasets information signal, where correct contains the highest amount of information, incorrect contains less but still non-zero due to the symmetry of the time series, and irrelevant and shuffles represent the null where the text contributes no additional information. 

\clearpage
\newpage
\section{WEATHER5K Full Results}
\label{sec:appendix-weather5k-results}
\begin{longtable}{lccccccccc}
\caption{WEATHER-5K, results using ridge forecaster model on the 300-station subset, reported as MSE per variable for every lag selector, candidate pool and budget. Random rows are means over \(n=20\) draws with the sd of the normalized metric. $\Delta$ is against the 48\,h default (240 features).} \label{tab:A_weather5k_ridge_MSE_full} \\
\toprule
\centering
\tiny
 &  & temp. & dew& wind& wind& pres- & MSE& sd & \(\Delta\)\\
inputs & n feat. &  &point & dir. & speed & sure &norm.& norm. &  (\%) \\
\midrule
\endfirsthead
\caption[]{WEATHER-5K, ridge forecaster on the 300-station subset, test MSE per variable for every lag selector, candidate pool and budget (same ridge, $\lambda$ rule and test rows throughout). Random rows are means over draws with the sd of the normalised metric. $\Delta$ is against the 48\,h default (240 features).} \\
\toprule
 &  & temp. & dew& wind& wind& pres- & MSE& sd &  \(\Delta\)\\
inputs & n feat. &  &point & dir. & speed & sure &norm.& norm. & (\%) \\
\midrule
\endhead
\midrule
\multicolumn{10}{r}{Continued on next page} \\
\midrule
\endfoot
\bottomrule
\endlastfoot
\multicolumn{6}{l}{\textit{matched (67)}}\\
\multirow[t]{4}{*}{\small Random} & 40 & 20.9 & 21.6 & 9086 & 5.04 & 53.8 & 0.507 & 0.008  & 1.17 \\
& 60 & 19.4 & 20.9 & 8956 & 4.87 & 52.4 & 0.494 & 0.006  & -1.51 \\
& 80 & 18.7 & 20.5 & 8869 & 4.79 & 51.4 & 0.486 & 0.006 & -3.07 \\
& 120 & 18.3 & 20.3 & 8800 & 4.71 & 51.3 & 0.481 & 0.004  & -3.97 \\
\midrule
\multirow[t]{4}{*}{\small ATEMPO} & 40 & 20.4 & 22.0 & 9056 & 5.08 & 52.9 & 0.506 & -- &  0.839 \\
& 60 & 19.7 & 21.4 & 8990 & 5.00 & 52.3 & 0.499 & -- & -0.483 \\
& 80 & 19.2 & 21.0 & 8939 & 4.93 & 51.9 & 0.494 & -- &  -1.51 \\
& 120 & 18.3 & 20.2 & 8813 & 4.75 & 50.7 & 0.481 & -- &  -3.99 \\
\midrule
\multirow[t]{4}{*}{\small TENDE} & 40 & 25.7 & 26.1 & 9247 & 5.20 & 64.3 & 0.552 & -- &  10.1 \\
& 60 & 24.8 & 25.9 & 9165 & 5.11 & 64.2 & 0.546 & -- & 8.88 \\
& 80 & 24.1 & 25.7 & 9117 & 5.05 & 64.0 & 0.542 & -- & 8.02 \\
& 120 & 23.3 & 25.3 & 9039 & 4.98 & 63.5 & 0.535 & -- & 6.78 \\
\midrule
\multirow[t]{4}{*}{MINE} & 40 & 23.3 & 22.8 & 9161 & 5.16 & 57.2 & 0.525 & -- & 4.63 \\
& 60 & 20.5 & 22.2 & 9066 & 4.98 & 56.5 & 0.512 & -- &  2.08 \\
& 80 & 20.1 & 21.9 & 9009 & 4.91 & 55.5 & 0.505 & -- &  0.762 \\
& 120 & 19.5 & 21.7 & 8929 & 4.83 & 55.3 & 0.500 & -- &  -0.235 \\
\midrule
\multirow[t]{4}{*}{CCMI} & 40 & 20.0 & 21.6 & 9022 & 5.02 & 52.3 & 0.501 &  -- & -0.0762 \\
& 60 & 19.6 & 21.4 & 8972 & 4.96 & 52.0 & 0.497 & -- &  -0.932 \\
& 80 & 18.9 & 20.9 & 8906 & 4.85 & 51.4 & 0.489 & -- &  -2.40 \\
& 120 & 18.1 & 20.1 & 8761 & 4.69 & 50.5 & 0.478 & -- &  -4.72 \\
\midrule
\multirow[t]{4}{*}{NJEE} & 40 & 20.1 & 21.1 & 9011 & 4.99 & 52.2 & 0.499 &  -- & -0.474 \\
& 60 & 19.2 & 20.8 & 8923 & 4.88 & 51.6 & 0.491 & -- & -2.03 \\
& 80 & 18.6 & 20.5 & 8852 & 4.79 & 51.1 & 0.485 & -- & -3.30 \\
& 120 & 18.0 & 20.0 & 8747 & 4.67 & 50.5 & 0.477 & -- &  -4.94 \\
\midrule
\multirow[t]{4}{*}{\small NPEET} & 40 & 22.2 & 21.2 & 9021 & 5.07 & 51.9 & 0.503 & -- & 0.368 \\
& 60 & 19.5 & 21.0 & 8935 & 4.91 & 51.7 & 0.493 & -- &  -1.68 \\
& 80 & 18.8 & 20.4 & 8864 & 4.80 & 50.8 & 0.485 & -- & -3.26 \\
& 120 & 18.3 & 20.3 & 8793 & 4.74 & 50.8 & 0.481 & -- & -4.10 \\
\midrule
\multirow[t]{4}{*}{PCMI} & 40 & 33.2 & 23.9 & 9276 & 5.59 & 59.4 & 0.558 &  -- & 11.2 \\
& 60 & 32.9 & 23.4 & 9222 & 5.52 & 58.8 & 0.552 & -- & 10.1 \\
& 80 & 22.2 & 20.7 & 8971 & 5.03 & 51.9 & 0.500 & -- & -0.202 \\
& 120 & 18.7 & 20.4 & 8841 & 4.77 & 51.3 & 0.484 & -- &  -3.36 \\
\midrule
\multirow[t]{4}{*}{\small \makecell[l]{\\Liang-\\Kleeman}}  & 40 & 26.6 & 25.1 & 9235 & 5.55 & 54.9 & 0.539 & -- &  7.42 \\
& 60 & 23.0 & 23.9 & 9138 & 5.31 & 54.1 & 0.522 & -- &  4.12 \\
& 80 & 21.7 & 23.7 & 9097 & 5.23 & 53.9 & 0.517 & -- &  3.14 \\
& 120 & 21.2 & 23.7 & 9062 & 5.21 & 54.0 & 0.515 & -- &  2.72 \\
\midrule
\multirow[t]{4}{*}{DTW} & 40 & 22.7 & 23.8 & 9131 & 5.30 & 54.1 & 0.521 & -- & 3.98 \\
& 60 & 21.3 & 23.7 & 9083 & 5.21 & 54.0 & 0.516 & -- &  2.86 \\
& 80 & 20.4 & 22.6 & 9016 & 5.08 & 53.0 & 0.506 & -- &  0.898 \\
& 120 & 19.6 & 21.8 & 8949 & 4.98 & 52.4 & 0.498 & -- &  -0.628 \\
\cline{2-10} \midrule
\multicolumn{6}{l}{\textit{full (336)}}\\
\makecell[l]{Default\\(48 h)} & 240 & 19.9 & 22.4 & 8939 & 5.01 & 53.0 & 0.501 & -- &  0.00 \\
\midrule
\multirow[t]{5}{*}{\small Random} & 40 & 23.0 & 23.5 & 9181 & 5.12 & 58.7 & 0.528 & 0.018 &  5.39 \\
 & 60 & 20.9 & 22.3 & 9063 & 4.96 & 56.1 & 0.511 & 0.017  & 1.95 \\
& 80 & 20.3 & 22.1 & 8995 & 4.89 & 55.9 & 0.506 & 0.016  & 0.943 \\
 & 120 & 19.5 & 21.5 & 8895 & 4.79 & 54.7 & 0.497 & 0.016  & -0.947 \\
& 240 & 18.3 & 20.6 & 8730 & 4.64 & 52.5 & 0.481 & 0.012  & -4.01 \\
\midrule
\multirow[t]{5}{*}{\makecell[l]{\\\\most\\recent\\n}} & 40 & 26.6 & 25.1 & 9235 & 5.55 & 54.9 & 0.539 & -- &  7.42 \\
& 60 & 23.3 & 24.5 & 9169 & 5.40 & 54.6 & 0.528 & -- & 5.23 \\
& 80 & 22.0 & 24.2 & 9124 & 5.30 & 54.3 & 0.521 & -- &  3.96 \\
& 120 & 21.2 & 23.7 & 9062 & 5.21 & 54.0 & 0.515 & -- & 2.72 \\
& 240 & 19.9 & 22.4 & 8939 & 5.01 & 53.0 & 0.501 & -- &  0.00 \\
\midrule
\multirow[t]{5}{*}{\small ATEMPO} & 40 & 22.9 & 22.1 & 9108 & 5.19 & 53.2 & 0.514 & -- &  2.45 \\
& 60 & 19.5 & 21.1 & 8986 & 4.96 & 51.8 & 0.496 & -- & -1.08 \\
& 80 & 19.1 & 20.9 & 8935 & 4.90 & 51.8 & 0.492 & -- & -1.81 \\
& 120 & 18.3 & 20.3 & 8825 & 4.76 & 51.0 & 0.482 & -- &  -3.76 \\
& 240 & 17.7 & 19.9 & 8684 & 4.62 & 50.4 & 0.473 & -- & -5.63 \\
\midrule
\multirow[t]{5}{*}{MINE} & 40 & 22.3 & 23.2 & 9186 & 5.12 & 58.6 & 0.527 & -- & 5.11 \\
& 60 & 21.2 & 22.8 & 9083 & 4.98 & 58.2 & 0.518 & -- &  3.31 \\
& 80 & 21.0 & 22.7 & 9042 & 4.94 & 58.1 & 0.515 & -- &  2.81 \\
& 120 & 20.4 & 22.6 & 8964 & 4.86 & 58.0 & 0.510 & -- &  1.82 \\
& 240 & 19.2 & 21.7 & 8809 & 4.72 & 55.8 & 0.496 & -- &  -1.15 \\
\midrule
\multirow[t]{5}{*}{CCMI} & 40 & 20.9 & 22.7 & 9069 & 5.14 & 53.1 & 0.510 &  -- & 1.63 \\
& 60 & 19.3 & 21.3 & 8968 & 4.95 & 51.8 & 0.495 & -- & -1.20 \\
& 80 & 18.6 & 20.5 & 8883 & 4.82 & 51.0 & 0.486 & -- &  -3.03 \\
& 120 & 18.1 & 20.2 & 8790 & 4.71 & 50.7 & 0.479 & -- &  -4.36 \\
 & 240 & 17.6 & 19.9 & 8646 & 4.58 & 50.3 & 0.471 & -- &  -6.10 \\
\midrule
\multirow[t]{5}{*}{NJEE} & 40 & 23.0 & 23.9 & 9224 & 5.17 & 59.0 & 0.531 &  -- & 6.01 \\
& 60 & 22.2 & 23.7 & 9163 & 5.08 & 58.8 & 0.526 & -- &  4.98 \\
 & 80 & 19.5 & 20.7 & 8919 & 4.81 & 51.6 & 0.489 & -- & -2.43 \\
& 120 & 18.8 & 20.4 & 8842 & 4.73 & 51.2 & 0.483 & -- &  -3.56 \\
& 240 & 18.0 & 20.0 & 8700 & 4.61 & 50.6 & 0.474 & -- &  -5.38 \\
\midrule
\multirow[t]{5}{*}{\small NPEET} & 40 & 22.9 & 23.3 & 9126 & 5.27 & 53.6 & 0.519 & -- & 3.47 \\
& 60 & 20.8 & 23.1 & 9052 & 5.14 & 53.4 & 0.510 & -- &  1.81 \\
& 80 & 20.1 & 22.2 & 8999 & 5.03 & 52.7 & 0.503 & -- & 0.238 \\
& 120 & 19.7 & 22.1 & 8950 & 4.98 & 52.6 & 0.499 & -- &  -0.427 \\
& 240 & 18.5 & 20.8 & 8808 & 4.78 & 51.4 & 0.485 & -- &  -3.30 \\
\midrule
\multirow[t]{5}{*}{PCMI} & 40 & 33.2 & 23.9 & 9276 & 5.59 & 59.4 & 0.558 &  -- & 11.2 \\
& 60 & 32.9 & 23.4 & 9222 & 5.52 & 58.8 & 0.552 & -- &10.1 \\
& 80 & 22.2 & 20.7 & 8971 & 5.03 & 51.9 & 0.500 & -- &  -0.202 \\
& 120 & 19.0 & 20.3 & 8814 & 4.77 & 51.3 & 0.484 & -- & -3.39 \\
& 240 & 18.0 & 20.1 & 8683 & 4.61 & 51.1 & 0.475 & -- &  -5.18 \\
\midrule
\multirow[t]{5}{*}{\small \makecell[l]{\\Liang-\\Kleeman}} & 40 & 26.6 & 25.1 & 9235 & 5.55 & 54.9 & 0.539 & -- &  7.42 \\
& 60 & 23.0 & 23.9 & 9138 & 5.31 & 54.1 & 0.522 & -- &  4.12 \\
& 80 & 21.7 & 23.7 & 9097 & 5.23 & 53.9 & 0.517 & -- &  3.14 \\
& 120 & 21.2 & 23.7 & 9062 & 5.21 & 54.0 & 0.515 & -- & 2.72 \\
& 240 & 18.8 & 20.8 & 8849 & 4.83 & 51.4 & 0.487 & -- &  -2.81 \\
\midrule
\multirow[t]{5}{*}{DTW} & 40 & 22.7 & 23.8 & 9131 & 5.30 & 54.1 & 0.521 &  -- & 3.98 \\
& 60 & 21.3 & 23.7 & 9083 & 5.21 & 54.0 & 0.516 & -- &  2.86 \\
& 80 & 20.4 & 22.6 & 9013 & 5.08 & 53.0 & 0.506 & -- &  0.876 \\
& 120 & 19.9 & 22.5 & 8958 & 5.01 & 53.0 & 0.502 & -- &  0.0922 \\
& 240 & 18.5 & 20.8 & 8795 & 4.78 & 51.5 & 0.485 & -- &  -3.32 \\
\cline{1-10} \cline{2-10} \midrule
\end{longtable}
\begin{table}[ht]
\centering
\tiny
\caption{WEATHER-5K, iTransformer, test MSE per variable in original units (pred\_len 168). `MSE norm.' is the benchmark's scaled-space metric; $\Delta$ is its \% change against the 48\,h default of the same station set.}
\label{tab:A_weather5k_itransformer_MSE_full}
\begin{tabular}{lllcccccccccc}
\toprule
  &  &  & steps & selection & train & temp.& dew & wind & wind & pressure & MSE & \(\Delta\)\\
 inputs & n feat. & seed & (x1000) & (h) & (min) &  & point & dir. & speed &  & norm. & (\%)  \\
\midrule
\multicolumn{6}{l}{\textit{(a) all 5{,}672 stations}}\\
 ATEMPO & 240 & 2021 & 300 & 1.787 & 79 & 18.373 & 21.828 & 8830.489 & 5.180 & 59.022 & 0.516 & -7.254 \\
\midrule
Default (48 h) & 240 & 2021 & 300 & 0.000 & 110 & 20.406 & 24.516 & 9374.630 & 5.613 & 63.687 & 0.556 & 0.000 \\
\midrule
\multicolumn{6}{l}{\textit{(b) 300 stations}}\\
\multirow[t]{5}{*}{ATEMPO} & \multirow[t]{5}{*}{240} & 2021 & 200 & 1.787 & 79 & 16.959 & 19.999 & 8672.705 & 4.559 & 58.924 & 0.491 & -5.951 \\
&  & 2022 & 200 & 1.787 & 49 & 17.063 & 20.065 & 8687.295 & 4.563 & 58.946 & 0.491 & -5.821 \\
&  & 2023 & 200 & 1.787 & 48 & 17.061 & 20.113 & 8682.039 & 4.558 & 59.163 & 0.492 & -5.759 \\
&  & 2024 & 200 & 1.787 & 49 & 17.055 & 20.050 & 8675.365 & 4.563 & 58.590 & 0.490 & -6.034 \\
&  & 2025 & 200 & 1.787 & 49 & 16.833 & 19.868 & 8670.379 & 4.552 & 58.052 & 0.488 & -6.454 \\
\midrule
\multirow[t]{5}{*}{CCMI} & \multirow[t]{5}{*}{240} & 2021 & 200 & 4.358 & 49 & 16.814 & 19.806 & 8664.002 & 4.543 & 59.308 & 0.491 & -5.976 \\
&  & 2022 & 200 & 4.358 & 49 & 16.771 & 19.753 & 8651.829 & 4.516 & 59.140 & 0.489 & -6.272 \\
&  & 2023 & 200 & 4.358 & 48 & 16.867 & 19.948 & 8654.402 & 4.542 & 59.362 & 0.491 & -5.946 \\
&  & 2024 & 200 & 4.358 & 49 & 16.833 & 19.764 & 8661.952 & 4.557 & 59.512 & 0.491 & -5.823 \\
&  & 2025 & 200 & 4.358 & 48 & 16.795 & 19.828 & 8655.864 & 4.540 & 59.342 & 0.490 & -6.007 \\
\midrule
\multirow[t]{3}{*}{DTW} & \multirow[t]{3}{*}{240} & 2021 & 200 & 0.186 & 59 & 17.875 & 21.107 & 8827.293 & 4.696 & 59.604 & 0.502 & -3.804 \\
&  & 2022 & 200 & 0.186 & 60 & 17.814 & 21.027 & 8820.951 & 4.698 & 60.140 & 0.503 & -3.609 \\
&  & 2023 & 200 & 0.186 & 61 & 17.751 & 20.996 & 8825.594 & 4.703 & 59.553 & 0.502 & -3.853 \\
\midrule
\multirow[t]{3}{*}{Liang} & \multirow[t]{3}{*}{240} & 2021 & 200 & 0.495 & 59 & 18.084 & 21.069 & 8892.234 & 4.736 & 58.896 & 0.503 & -3.623 \\
 &  & 2022 & 200 & 0.495 & 63 & 18.075 & 21.110 & 8906.986 & 4.730 & 58.955 & 0.503 & -3.562 \\
&  & 2023 & 200 & 0.495 & 62 & 18.040 & 21.019 & 8896.114 & 4.738 & 60.129 & 0.506 & -3.061 \\
\midrule
\multirow[t]{3}{*}{MINE} & \multirow[t]{3}{*}{240} & 2021 & 200 & 4.632 & 49 & 18.444 & 21.784 & 8876.295 & 4.722 & 66.527 & 0.522 & -0.017 \\
&  & 2022 & 200 & 4.632 & 51 & 18.497 & 21.973 & 8877.440 & 4.722 & 66.122 & 0.521 & -0.141 \\
&  & 2023 & 200 & 4.632 & 50 & 18.521 & 21.897 & 8895.313 & 4.725 & 67.203 & 0.524 & 0.422 \\
\midrule
\multirow[t]{3}{*}{NJEE} & \multirow[t]{3}{*}{240} & 2021 & 200 & 2.942 & 48 & 17.378 & 20.193 & 8705.982 & 4.586 & 60.687 & 0.497 & -4.729 \\
 &  & 2022 & 200 & 2.942 & 50 & 17.304 & 20.090 & 8706.281 & 4.576 & 60.268 & 0.496 & -5.014 \\
&  & 2023 & 200 & 2.942 & 50 & 17.303 & 20.190 & 8709.997 & 4.579 & 61.502 & 0.499 & -4.395 \\
\midrule
\multirow[t]{3}{*}{NPEET} & \multirow[t]{3}{*}{240} & 2021 & 200 & 2.596 & 59 & 17.817 & 20.972 & 8849.000 & 4.693 & 59.042 & 0.501 & -4.043 \\
&  & 2022 & 200 & 2.596 & 60 & 17.797 & 20.930 & 8848.920 & 4.693 & 59.201 & 0.501 & -3.987 \\
&  & 2023 & 200 & 2.596 & 61 & 17.751 & 20.910 & 8843.000 & 4.702 & 58.815 & 0.500 & -4.153 \\
\midrule
\multirow[t]{3}{*}{PCMI} & \multirow[t]{3}{*}{240} & 2021 & 200 & 1.289 & 48 & 17.212 & 20.229 & 8717.397 & 4.640 & 60.391 & 0.498 & -4.555 \\
&  & 2022 & 200 & 1.289 & 50 & 17.258 & 20.356 & 8726.672 & 4.625 & 60.597 & 0.498 & -4.459 \\
&  & 2023 & 200 & 1.289 & 50 & 17.344 & 20.377 & 8732.143 & 4.657 & 61.495 & 0.502 & -3.835 \\
\midrule
\multirow[t]{3}{*}{Random} & \multirow[t]{3}{*}{240} & 2021 & 200 & 0.000 & 73 & 17.680 & 20.812 & 8802.316 & 4.647 & 61.911 & 0.505 & -3.238 \\
&  & 2022 & 200 & 0.000 & 48 & 17.608 & 20.819 & 8805.193 & 4.656 & 62.792 & 0.507 & -2.794 \\
&  & 2023 & 200 & 0.000 & 49 & 17.719 & 20.941 & 8798.976 & 4.655 & 62.997 & 0.508 & -2.671 \\
\midrule
\multirow[t]{3}{*}{Default (48 h)} & \multirow[t]{3}{*}{240} & 2021 & 200 & 0.000 & 110 & 18.962 & 22.596 & 9175.466 & 4.903 & 61.157 & 0.522 & 0.000 \\
&  & 2022 & 200 & 0.000 & 94 & 18.937 & 22.573 & 9180.363 & 4.904 & 61.258 & 0.522 & 0.055 \\
&  & 2023 & 200 & 0.000 & 108 & 19.060 & 22.643 & 9171.314 & 4.911 & 61.582 & 0.523 & 0.252 \\
\midrule
\multirow[t]{5}{*}{ATEMPO} & \multirow[t]{5}{*}{120} & 2021 & 100 & 1.787 & 48 & 17.189 & 20.334 & 8925.412 & 4.697 & 56.133 & 0.494 & -5.358 \\
 &  & 2022 & 200 & 1.787 & 48 & 17.442 & 20.549 & 8930.164 & 4.697 & 58.555 & 0.500 & -4.123 \\
&  & 2023 & 200 & 1.787 & 57 & 17.441 & 20.437 & 8914.127 & 4.698 & 59.371 & 0.502 & -3.836 \\
 &  & 2024 & 200 & 1.787 & 49 & 17.384 & 20.415 & 8918.404 & 4.701 & 56.767 & 0.496 & -5.012 \\
&  & 2025 & 200 & 1.787 & 48 & 17.291 & 20.446 & 8922.896 & 4.707 & 58.760 & 0.501 & -4.062 \\
\midrule
\multirow[t]{5}{*}{CCMI} & \multirow[t]{5}{*}{120} & 2021 & 200 & 4.358 & 48 & 17.297 & 20.298 & 8846.284 & 4.649 & 56.961 & 0.493 & -5.531 \\
&  & 2022 & 200 & 4.358 & 49 & 17.377 & 20.355 & 8852.358 & 4.660 & 57.790 & 0.495 & -5.039 \\
 &  & 2023 & 200 & 4.358 & 48 & 17.358 & 20.394 & 8846.122 & 4.659 & 57.696 & 0.495 & -5.102 \\
&  & 2024 & 200 & 4.358 & 49 & 17.270 & 20.330 & 8847.415 & 4.660 & 57.626 & 0.495 & -5.165 \\
 &  & 2025 & 200 & 4.358 & 48 & 17.383 & 20.458 & 8852.969 & 4.658 & 57.779 & 0.496 & -5.025 \\
\midrule
\multirow[t]{3}{*}{MINE} & \multirow[t]{3}{*}{120} & 2021 & 200 & 4.632 & 49 & 19.693 & 23.010 & 9078.167 & 4.870 & 69.391 & 0.540 & 3.466 \\
&  & 2022 & 200 & 4.632 & 49 & 19.664 & 22.955 & 9069.425 & 4.867 & 69.517 & 0.540 & 3.453 \\
&  & 2023 & 200 & 4.632 & 48 & 19.692 & 22.889 & 9071.364 & 4.859 & 69.343 & 0.539 & 3.326 \\
\midrule
\multirow[t]{3}{*}{NJEE} & \multirow[t]{3}{*}{120} & 2021 & 200 & 2.942 & 47 & 18.055 & 20.730 & 8913.436 & 4.716 & 61.334 & 0.508 & -2.637 \\
&  & 2022 & 200 & 2.942 & 48 & 18.029 & 20.828 & 8915.019 & 4.703 & 61.620 & 0.508 & -2.552 \\
&  & 2023 & 200 & 2.942 & 47 & 17.845 & 20.625 & 8911.631 & 4.715 & 62.101 & 0.509 & -2.370 \\
\midrule
\multirow[t]{3}{*}{NPEET} & \multirow[t]{3}{*}{120} & 2021 & 200 & 2.596 & 77 & 19.058 & 22.648 & 9279.161 & 4.963 & 61.940 & 0.528 & 1.117 \\
&  & 2022 & 200 & 2.596 & 78 & 19.074 & 22.642 & 9286.872 & 4.968 & 61.033 & 0.526 & 0.765 \\
&  & 2023 & 200 & 2.596 & 89 & 19.067 & 22.656 & 9286.169 & 4.964 & 61.500 & 0.527 & 0.950 \\
\midrule
\multirow[t]{3}{*}{PCMI} & \multirow[t]{3}{*}{120} & 2021 & 200 & 1.289 & 48 & 18.199 & 20.716 & 8905.024 & 4.780 & 59.776 & 0.506 & -2.999 \\
&  & 2022 & 200 & 1.289 & 49 & 18.222 & 20.763 & 8906.501 & 4.773 & 59.921 & 0.506 & -2.951 \\
&  & 2023 & 200 & 1.289 & 48 & 18.204 & 20.735 & 8902.049 & 4.785 & 60.007 & 0.507 & -2.874 \\
\midrule
\multirow[t]{3}{*}{Random} & \multirow[t]{3}{*}{120} & 2021 & 100 & 0.000 & 47 & 18.357 & 21.364 & 8992.654 & 4.765 & 62.660 & 0.515 & -1.227 \\
&  & 2022 & 200 & 0.000 & 49 & 18.596 & 21.561 & 9003.798 & 4.776 & 63.630 & 0.519 & -0.578 \\
&  & 2023 & 200 & 0.000 & 47 & 18.511 & 21.550 & 8995.024 & 4.802 & 63.530 & 0.519 & -0.537 \\
\midrule
\multirow[t]{3}{*}{Default (24 h)} & \multirow[t]{3}{*}{120} & 2021 & 200 & 0.000 & 166 & 21.082 & 24.930 & 10021.410 & 5.373 & 65.506 & 0.568 & 8.878 \\
 &  & 2022 & 200 & 0.000 & 112 & 21.108 & 24.974 & 10031.114 & 5.377 & 65.483 & 0.568 & 8.943 \\
 &  & 2023 & 200 & 0.000 & 110 & 21.073 & 24.932 & 10020.902 & 5.367 & 65.454 & 0.568 & 8.818 \\
\midrule
\bottomrule
\end{tabular}
\end{table}
We follow the WEATHER-5K benchmark and report MSE in the original units across five features. We use two different forecasting model architectures, iTransformer as our main benchmark comparison in Table~\ref{tab:A_weather5k_itransformer_MSE_full}, and a ridge regression in Table~\ref{tab:A_weather5k_ridge_MSE_full}. 
%Our methodology implementation described in Section~\ref{sec:method} can be found at: \url{https://github.com/EmmaAndrews13/TENDE-with-lag}.

Using the simpler linear ridge model on WEATHER5K 300-station subset enables a larger number of experiments which would not have been possible using the more computationally expensive iTransformer architecture. Table~\ref{tab:A_weather5k_ridge_MSE_full} shows that despite its simplicity, the ridge regression model performs well. ATEMPO's improvements of \(~3.8\%\) between the 240 feature WEATHER5K baseline recent(48h) and the 120 feature ATEMPO features remains comparable to the \(~4.6\%\) improvement using iTransformer, showing the improvement from ATEMPO selected lags is robust across models. 

Further baselines were run, however TENDE was computationally intractable to run even across a 300 station subset, so instead we feed them a small sub-selected number of candidate lags, \(n=67\), designed in a grid which is dense at short lags and logarithmically sparser at long ones. Because a 67-candidate search is strictly easier than a 336-candidate one, we ran all baselines across both for a fair comparison. 
\clearpage
\newpage
\subsection{Physical analysis of selected lags}
\label{sec:appendix-physical-analysis}
\begin{table}
\centering
\small
\caption{Where each selector spends its feature budget: the lag sets the iTransformer arms were trained on, counted by band. Default refers to the most recent \(n=\)budget lags.}
\label{tab:physical-analysis}
\begin{tabular}{llcccccccc}
\toprule
 &  & persis & sub- & diurnal & synoptic & weekly & boundary & phase & median\\
budget & selector & -tence & daily &  &  &  &  & R &  lag (h) \\
\midrule
\multirow[t]{8}{*}{24} & Default & 3 & 19 & 2 & 0 & 0 & 0 & 0.00 & 12.50 \\
& Random & 0 & 2 & 0 & 21 & 1 & 0 & 0.29 & 162.00 \\
 & PCMI & 1 & 3 & 13 & 6 & 0 & 1 & 0.35 & 114.00 \\
 & NPEET & 2 & 13 & 1 & 8 & 0 & 0 & 0.11 & 17.50 \\
 & MINE & 0 & 0 & 2 & 20 & 2 & 0 & 0.20 & 154.50 \\
 & NJEE & 1 & 0 & 3 & 18 & 0 & 2 & 0.11 & 254.50 \\
& ATEMPO & 2 & 7 & 3 & 5 & 1 & 6 & 0.26 & 54.00 \\
 & CCMI & 2 & 9 & 2 & 11 & 0 & 0 & 0.18 & 40.00 \\
\cline{1-10}
\multirow[t]{10}{*}{48} & Default & 3 & 19 & 5 & 21 & 0 & 0 & 0.00 & 24.50 \\
 & Random & 0 & 4 & 4 & 35 & 3 & 2 & 0.19 & 147.50 \\
 & \makecell[l]{Liang-\\Kleeman} & 3 & 19 & 12 & 14 & 0 & 0 & 0.32 & 24.50 \\
  & DTW & 3 & 19 & 7 & 19 & 0 & 0 & 0.11 & 24.50 \\
 & PCMI & 1 & 5 & 13 & 27 & 1 & 1 & 0.23 & 136.50 \\
 & NPEET & 3 & 16 & 5 & 24 & 0 & 0 & 0.10 & 28.50 \\
 & MINE & 0 & 1 & 5 & 40 & 2 & 0 & 0.07 & 170.00 \\
 & NJEE & 2 & 0 & 4 & 36 & 1 & 5 & 0.11 & 241.00 \\
 & CCMI & 2 & 11 & 7 & 28 & 0 & 0 & 0.12 & 111.00 \\
  & ATEMPO & 3 & 7 & 11 & 19 & 1 & 7 & 0.30 & 118.00 \\
\cline{1-10}
\bottomrule
\end{tabular}
\end{table}
\begin{table}
\centering
\small
\caption{Which of ATEMPO's selected lags do the forecasting work. Ridge on 300 stations, scored using mean MSE over variables. Each band row removes that band from ATEMPO's set and reports the \% change of the metric relative to the full set: a positive \emph{cost of band} means the forecast got worse without it, i.e. the band was helping; \emph{per lag (band)} divides by the band's size, and the leave-one-out (LOO) columns give the mean and the largest change from removing a single lag of the band.
}
\label{tab:A_weather5k_band_ablation_mse}
\begin{tabular}{llrrrrr}
\toprule
 &  & $n$ lags  & cost of & per lag & per lag & per lag \\
budget & band &   & band (\%) & (band) & (LOO mean)  & (LOO max) \\
\midrule
\multirow[t]{6}{*}{24} 
 & persistence & 2  & 1.54 & 0.77 & 0.30 & 0.55 \\
 & sub-daily & 7 & 1.42 & 0.20 & 0.05 & 0.11 \\
 & diurnal & 3  & 0.43 & 0.14 & 0.11 & 0.12 \\
 & synoptic & 4  & 1.18 & 0.30 & 0.20 & 0.33 \\
 & weekly & 1  & 0.86 & 0.86 & 0.86 & 0.86 \\
 & boundary & 7 & 0.28 & 0.04 & 0.01 & 0.04 \\
\cline{1-7}
\multirow[t]{6}{*}{48}
 & persistence & 3  & 1.44 & 0.48 & 0.18 & 0.52 \\
 & sub-daily & 7 & 0.66 & 0.09 & 0.02 & 0.06 \\
 & diurnal & 11  & 0.59 & 0.05 & 0.03 & 0.06 \\
 & synoptic & 18  & 2.19 & 0.12 & 0.04 & 0.14 \\
 & weekly & 1 & 0.14 & 0.14 & 0.14 & 0.14 \\
 & boundary & 8  & 0.09 & 0.01 & 0.00 & 0.02 \\
\cline{1-7}
\bottomrule
\end{tabular}
\end{table}

WEATHER5K is not labelled with ground truth lags, however weather data does have natural patterns we can use to compare the selected lags across baselines. Specifically, we classify each selector's lags into persistence (the most recent 1-3 hours), sub-daily (within 24 hours), diurnal harmonics (multiples of 24hours, \(\pm1\)), synoptic (everything else between 1-14 days), weekly, and boundary band (the final seven hours 330-336). Phase $R$ is the circular concentration of the lags on the 24 h clock (1 = every lag at the same time of day), and random is the single draw the random arm used. We used the 24-lag sets that were used to train an iTransformer model, with full forecasting results available in Appendix~\ref{sec:appendix-weather5k-results}.

Table~\ref{tab:physical-analysis} shows the results. NPEET only selects lags occuring in the first 59 hours. PCMI over-selects lags that are all spaced 24 hours apart, which can make physical sense if a certain hour of day is the strongest driving signal, but also likely increases the amount of redundant information contained in the lag sub-set. MINE and NJEE forecasting performance are indistinguishable from random and this is supported as most of their lags are not well distributed and have no obvious pattern apart from being in the first 1-14 days. ATEMPO is the only estimator that spreads lags across all our categories, reflecting it's strong ability to select lags that maximize information for downstream forecasting.

However ATEMPO does select a surprisingly high number of lags in the final 6 hours, where all other estimators select zero. We perform an ablation that drops each lag band in turn from ATEMPO's chosen lags to see the effect on forecasting performance using the ridge model. Table~\ref{tab:A_weather5k_band_ablation_mse} shows the change in MSE when removing a lag from a specific band. \emph{cost of band} is the change in $\Delta$ from removing it, so positive means the set got worse and the band was helping. The leave-one-out (LOO) columns give the mean and the largest cost of removing a single lag of the band. The boundary lags are consistently the least valuable, with the smallest positive cost across all three columns. Removing any band removes information, so we would always expect a small positive change.

Our hypothesis is the high number of boundary layer lags may be an artifact of how the delay profile is calculated when \(k \rightarrow K\). It can be understood through the geometry of the lag encoding; TE is estimated from Eq.~\ref{eq:final}, whose inner term is the squared score difference between the two conditionings. These differ in exactly \(k\) coordinates and therefore have a Euclidean distance of \(\sqrt{k}\), resulting in approximation errors that de-correlate as \(k\) increases. This can produce a residual error that also grows with \(k\). This artefact would then create the largest effect at large \(k\), corresponding the boundary errors observed as \(k \rightarrow K\). 
% \section{Amortization Artefacts}
% The marginal gain, \(\Delta TE\) increases sharply towards \(k=K_{\text{max}}\). Fitting TENDE for each \((k,l)\) pair individually, with no parameters shared across cells, lowers this effect from a median ramp of \(+4.44\) to \(0.08\) on Weather5K, meaning it is an artefact of amoritization rather than the data. It can be understood through the geometry of the lag encoding; TE is estimated from Equation~\ref{eq:estimator}, whose iner term is the squared score difference \(\left[\tfrac12\left\lVert \epsilon^{(k,l)} - \epsilon^{(0,l)} \right\rVert^2\right]\), where \(\epsilon^{(k,l)}\) is the amortized network's output when conditioned on the encoding \(\text{enc}(k,l)\). The encoding \(\text{enc}(k,l)\) is at Euclidean distance \(\sqrt{k}\) from the marginalized encoding \(\text{enc}(0,l)\), as they differ in exactly \(k\) coordinates. This results in approximation errors that decorrelate as \(k\) increases, meaning they do not cancel and instead the difference is squared, producing a residual error that also grows with \(k\). Replacing the increasing \(k\) with encodings held at a unit distance apart for every \(k\) removes the ramp on our synthetic data with known ground truth from \(2.00\) to \(-0.01\), confirming it is a property of encoding geometry rather than amortization itself. 

\section{Additional Models}
\label{sec:appendix-additional-models}
The paper's WEATHER-5K result is that a forecaster given 48 hours chosen by ATEMPO from a 336-hour history beats the same
forecaster given the most recent 48 hours, at the same input size, and that the margin tracks the quality of the lag
selector. Every arm of that result was trained with iTransformer, chosen because its inverted embedding treats the input
rows as a feature vector and so accepts non-consecutive hours without modification. Here we answer three more questions:
\begin{itemize}
    \item Does the ranking of lag sets survive a change of architecture?
    \item Does it matter whether the selected hours are
handed to the model as a short gathered sequence or as a long contiguous window with the unselected hours blanked?
    \item Does a pretrained foundation model, which brings its own prior over hourly dynamics, see the same gain from the selected
hours? 
\end{itemize}

We test these questions over input choices at the 48-lag budget: the benchmark's default window (the
48 most recent hours), 48 random hours of the 336, the 48 hours ranked top by ATEMPO, and the 48 ranked top by CCMI (the strongest competing selector in the paper). The full 336-hour window is the reference every selection is trying to approximate at one seventh of the input size.

\subsection{Model Details}
iTransformer and DLinear take input rows unchanged. PatchTST embeds patches of 16 adjacent rows with stride 8 and Pyraformer builds a pyramid of coarser scales with convolutions over adjacent rows, so both carry an adjacency assumption that is not true for our selected lag input. All four train with the benchmark parameters stated in Section~\ref{sec:appendix-dataset-implementations} (300-station subset, batch 1024, 200k-step ceiling, validation every 20k steps, patience 3, cosine schedule, $d_{\mathrm{model}}=512$, three encoder layers), apart from Pyraformer. Due to compute constraints, Pyraformer on 336-hour inputs keeps a $(\text{batch}, 8, 441, 441)$ attention tensor per layer for the backward pass (74\,MB per sample in fp32), so it runs batch 256 under mixed precision with a four-fold step ceiling and validation interval, which keeps the number of samples seen, the validation points in sample space and the learning-rate schedule the same. Chronos-2 (120M parameters) is evaluated zero-shot, forecasting the five variables of a station jointly from the same context, with the benchmark's own test loader and metric code. Chronos-2 assumes evenly spaced inputs and cannot take gathered rows. All experiments run across three seeds except for Pyraformer, the 336-hour PatchTST cells and Chronos-2 due to compute constraints. The seed standard deviation of the normalized MAE is below 0.001 in every multi-seed cell.
\begin{table*}[t]
\centering
\caption{WEATHER-5K at the benchmark's input size (48 selected lags, 300 stations) for five forecaster families, testing whether the ordering of Table~\ref{tab:weather5k} (default $>$ random $>$ the two strongest selectors) depends on the forecaster or on the input layout. MSE is averaged over the five variables and over training seeds ($\pm$ sd across seeds where more than one; $<$0.0001 where smaller); $\Delta$ is the percentage change against the default of the same panel. The check row marks whether the ordering holds. The last row gives every model the full 336\,h window, $\Delta$ against the default.}
%($\downarrow$) & \makecell{$\Delta$ vs \\default ($\downarrow$)}
\label{tab:A_weather5k_models}
\resizebox{\textwidth}{!}{%
\begin{tabular}{l|cc|cc|cc|cc|cc}
\toprule
 & \multicolumn{2}{c|}{iTransformer} & \multicolumn{2}{c|}{DLinear} & \multicolumn{2}{c|}{PatchTST} & \multicolumn{2}{c|}{Pyraformer} & \multicolumn{2}{c}{Chronos-2$^\dagger$} \\
 & MSE ($\downarrow$) & $\Delta (\downarrow)$ & MSE ($\downarrow$) & $\Delta (\downarrow)$ & MSE ($\downarrow$) & $\Delta (\downarrow)$ & MSE ($\downarrow$) & $\Delta (\downarrow)$ & MSE ($\downarrow$) & $\Delta (\downarrow)$ \\
\midrule
Default & 0.5223\,{\scriptsize$\pm$0.0007} & - & 0.5510\,{\scriptsize$<$0.0001} & - & 0.5480\,{\scriptsize$<$0.0001} & - & 0.4689 & - & 0.6478 & - \\
Random & 0.5066\,{\scriptsize$\pm$0.0016} & -3.00\% & 0.5040\,{\scriptsize$<$0.0001} & -8.53\% & 0.5004\,{\scriptsize$\pm$0.0002} & -8.69\% & 0.4901 & +4.52\% & 0.6351 & -1.96\% \\
CCMI & 0.4904\,{\scriptsize$\pm$0.0009} & -6.10\% & 0.4917\,{\scriptsize$<$0.0001} & -10.77\% & 0.4859\,{\scriptsize$\pm$0.0006} & -11.33\% & 0.4748 & +1.26\% & 0.5959 & -8.01\% \\
ATEMPO & 0.4904\,{\scriptsize$\pm$0.0014} & -6.10\% & 0.4948\,{\scriptsize$<$0.0001} & -10.21\% & 0.4879\,{\scriptsize$\pm$0.0009} & -10.97\% & 0.4774 & +1.82\% & 0.6239 & -3.69\% \\
\makecell[l]{\textit{Default $>$ Random}\\ \textit{$>$ CCMI, ATEMPO}} & \multicolumn{2}{c|}{\checkmark} & \multicolumn{2}{c|}{\checkmark} & \multicolumn{2}{c|}{\checkmark} & \multicolumn{2}{c|}{$\times$ (default best)} & \multicolumn{2}{c}{\checkmark} \\
\midrule
\textit{Full window (336\,h)} & 0.4918\,{\scriptsize$\pm$0.0018} & -5.84\% & 0.4821\,{\scriptsize$<$0.0001} & -12.50\% & 0.4962 & -9.45\% & 0.4679 & -0.22\% & 0.5312 & -18.00\% \\
\bottomrule
\end{tabular}}
\par\smallskip\footnotesize $^\dagger$ Chronos-2 is evaluated zero-shot (\texttt{amazon/chronos-2} and assumes evenly spaced inputs, so we pass the full a contiguous 336\,h window in which every hour outside the lag set is marked missing.
\end{table*}
\begin{table}[t]
\centering
\small
\caption{Every lag set of the paper's grid under the two architectures that take gathered inputs unchanged. Normalized MSE, mean over seeds, and \% change against each model's own 48\,h default (300 stations). Spearman rank correlation of the two columns across selectors: 0.82 at 48 lags ($n=10$), 0.88 at 24 lags ($n=8$).}
\label{tab:grid}
\begin{tabular}{llrrrrrr}
\toprule
 & & \multicolumn{3}{c}{iTransformer} & \multicolumn{3}{c}{DLinear} \\
\cmidrule(lr){3-5} \cmidrule(lr){6-8}
lags & selector & seeds & MSE norm. & $\Delta$ (\%) & seeds & MSE norm. & $\Delta$ (\%) \\
\midrule
48 & Default & 3 & 0.522 & 0.0 & 3 & 0.551 & 0.0 \\
 & Random & 3 & 0.507 & -3.0 & 3 & 0.504 & -8.5 \\
 & ATEMPO & 5 & 0.490 & -6.1 & 3 & 0.495 & -10.2 \\
 & CCMI & 5 & 0.490 & -6.1 & 3 & 0.492 & -10.8 \\
 & NJEE & 3 & 0.497 & -4.8 & 3 & 0.497 & -9.9 \\
 & NPEET & 3 & 0.501 & -4.2 & 3 & 0.519 & -5.8 \\
 & PCMI & 3 & 0.499 & -4.4 & 3 & 0.498 & -9.6 \\
 & MINE & 3 & 0.522 & -0.0 & 3 & 0.515 & -6.6 \\
 & Liang & 3 & 0.504 & -3.5 & 3 & 0.524 & -4.8 \\
 & DTW & 3 & 0.502 & -3.9 & 3 & 0.520 & -5.7 \\
\addlinespace[3pt]
24 & Default & 3 & 0.568 & 8.8 & 3 & 0.579 & 5.0 \\
 & Random & 2 & 0.519 & -0.7 & 3 & 0.520 & -5.7 \\
 & ATEMPO & 4 & 0.500 & -4.4 & 3 & 0.512 & -7.1 \\
 & CCMI & 5 & 0.495 & -5.3 & 3 & 0.507 & -7.9 \\
 & NJEE & 3 & 0.509 & -2.6 & 3 & 0.511 & -7.2 \\
 & NPEET & 3 & 0.527 & 0.8 & 3 & 0.547 & -0.7 \\
 & PCMI & 3 & 0.506 & -3.0 & 3 & 0.512 & -7.1 \\
 & MINE & 3 & 0.540 & 3.3 & 3 & 0.533 & -3.2 \\
\bottomrule
\end{tabular}
\end{table}
\paragraph{The ranking of lag sets does not depend on the forecaster.} Table~\ref{tab:A_weather5k_models} shows under every trained architecture except Pyraformer, the order is the same as in the main papers Table~\ref{tab:weather5k}: the default window is worst, random hours sit in the middle, and the ATEMPO and CCMI sets are best and close to each other. Table~\ref{tab:grid} extends this to every baseline at both budgets of 24 and 48 lags for the two architectures that take gathered rows unchanged, iTransformer and DLinear. The Spearman correlation between the iTransformer and DLinear columns is 0.82 at 48 lags and 0.88 at 24, so the paper's ordering of selectors is a property of the lag sets, not of the transformer that scored them.
%Pyraformer's contiguous 48\,h default performing best can be explained by its pyramidal attention and the convolutional bottleneck that builds the coarser levels both assume adjacent inputs one hour apart, so a gathered row of non-adjacent hours is a distorted input for it. Therefore a model built around multiscale structure of a contiguous series gains little from being told which hours matter.

This supports our conclusion that a subset of selected hours match or beat the full window across a range of forecaster models. For iTransformer and PatchTST the selected hours are slightly better than the whole history at one seventh of the input size, consistent with the extra inputs adding variance faster than they add signal. The exception is the zero-shot
model: Chronos-2 is pretrained to read long contexts and gains 18.0\% from the full window, far more than from
any 48-hour selection.

\section{NLB Full Results}
\label{sec:appendix-neural}
%
% \begin{table}[t]
% \centering\small
% \caption{The benchmark's own task on the NLB session: decoding current finger velocity from all 130 smoothed units with a ridge regression, no own-past input, val $R^2$. The current neural bin (lag 0) is in every decoder's base ($R^2$ 0.220); the selectors choose $B$ further lags from 20 to 380 ms of lead. Reference: the benchmark-style single fixed lag is best at 140 ms (0.397); all 20 lags 0.479.}
% \label{tab:nlb_neural}
% \begin{tabular}{lrrrr}
% \toprule
%  & $+1$ & $+2$ & $+3$ & $+5$ \\
% additional lags chosen by &  &  &  &  \\
% \midrule
% ATEMPO & 0.246 & 0.305 & 0.336 & \textbf{0.474} \\
% cross-correlation & \textbf{0.382} & \textbf{0.400} & \textbf{0.425} & 0.448 \\
% most recent lags & 0.246 & 0.305 & 0.336 & 0.402 \\
% evenly spaced & 0.246 & 0.267 & 0.384 & 0.468 \\
% % greedy wrapper (ridge refits) & 0.393 & 0.447 & 0.472 & 0.487 \\
% \bottomrule
% \end{tabular}
% \end{table}
\begin{table}[t]
\centering
\scriptsize
\caption{Lag budgets on the three NLB tasks. For NLB velocity and decoding, forecasting input is lag 0 plus $B$ selected input lags. For MC\_Maze there is no base input, only $B$ selected lag bins. $R^2$ calculated on held-out validation split.}
\label{tab:nlb_neural}
\begin{tabular}{lrrrr|rrrr|rrrr}
\toprule
 additional lags & \multicolumn{4}{c|}{NLB velocity} & \multicolumn{4}{c|}{NLB decoding} & \multicolumn{4}{c}{MC\_Maze decoding} \\
chosen by & $+1$ & $+2$ & $+3$ & $+5$ & $+1$ & $+2$ & $+3$ & $+5$ & $+1$ & $+2$ & $+3$ & $+5$ \\
\midrule
ATEMPO & 0.699 & 0.728 & 0.755 & \textbf{0.812} & 0.246 & 0.305 & 0.336 & \textbf{0.474} & 0.356 & \textbf{0.518} & \textbf{0.557} & 0.619 \\
cross-correlation & \textbf{0.727} & \textbf{0.754} & 0.755 & 0.794 & \textbf{0.382} & \textbf{0.400} & \textbf{0.425} & 0.448 & \textbf{0.359} & 0.395 & 0.449 & 0.504 \\
most recent lags & 0.699 & 0.728 & 0.755 & 0.794 & 0.246 & 0.305 & 0.336 & 0.402 & 0.356 & 0.394 & 0.453 & 0.504 \\
evenly spaced & 0.699 & 0.699 & 0.699 & 0.792 & 0.246 & 0.267 & 0.384 & 0.468 & 0.356 & 0.441 & 0.552 & \textbf{0.626} \\
random & 0.717 & 0.737 & \textbf{0.762} & 0.790 & 0.345 & 0.375 & 0.415 & 0.448 & 0.332 & 0.432 & 0.515 & 0.580 \\
\bottomrule
\end{tabular}
\end{table}
We test ATEMPO across three lag selection tasks from Neural Latents Benchmark~\cite{PeiYe2021NeuralLatents}. Table~\ref{tab:nlb_neural} reports validation $R^2$ across lag budgets, and compares ATEMPO to recent, which keeps the $B$ most recent lags (the field's default), cross-correlation, which keeps the $B$ lags with the largest absolute training-set correlation with the target, and random, which keeps $B$ lags drawn at random (mean over 20 draws). We quantify uncertainty with block-bootstrap of the held-out data: for each task, the held-out $R^2$ (per-dimension, averaged over $v_x$ and $v_y$) is recomputed on 2{,}000 bootstrap resamples of the evaluation set. Per-cell standard errors are $\le 0.016$ for NLB velocity, $\le 0.026$ for NLB decoding and
$\le 0.014$ for MC\_Maze. These overstate the uncertainty of \emph{comparisons}, because every method is scored on the same held-out blocks and a hard block lowers all methods together. We therefore also bootstrap the paired differences, computing each method's $R^2$ and ATEMPO's on the same resamples: the paired standard errors are far smaller, at most $0.003$, $0.010$ and $0.011$ for the three tasks respectively. By this paired test, ATEMPO's advantage in Table~\ref{tab:te-estimate} over the next-best estimator exceeds paired standard errors on both decoding tasks, and on NLB velocity, CCMI's $+0.001$ is below one paired standard error - a statistical tie.

%These tasks involve physiologically established delays, providing a basis for evaluating lag selection. In NLB velocity, we predict a monkey's finger velocity 200,ms ahead using its past velocity and selected lags of the target-position error, which affects movement after a visumotor feedback delay. In NLB decoding, we predict current finger velocity from neural spiking activity, comparing lag selection against the benchmark's standard assumption of a fixed 140,ms neural lead. In MC\_Maze decoding, we predict hand velocity at movement onset using only preparatory neural activity, making accurate neural lead-time selection essential.

Table~\ref{tab:nlb_neural} shows further forecasting results across $B$ selected lags for all three tasks. NLB velocity forecasts finger velocity 200\,ms ahead from its own past and the target-position
error the animal is correcting, whose influence on movement is delayed by the visuomotor feedback
latency. The current bin (lag 0) alone gives $R^2 = 0.67$ and all 30 input lags 0.82. At one to three additional lags the cumulative ATEMPO ranking selects the shallowest informative lags and therefore coincides
with the most-recent-lags baseline. The rankings separate at five, where ATEMPO's deeper picks (120 and 160\,ms, i.e.\ 320--360\,ms before the forecast target)
reach $0.812$ against $0.794$ for both recent lags and cross-correlation.

NLB decoding decodes current finger velocity from population rate. For reference we sweep a single fixed lead 0--380\,ms and find it peaks at a lead of 140\,ms, which is the value used by the benchmark to pair behavior with neural activity. This known lead gives $R^2 = 0.40$ compared to 0.22 for the current bin (lag 0) alone and 0.48 for all 20 lags. The current bin plus ATEMPO's top five lags reaches 0.47, close to the all-lags ceiling with less than a third of the inputs. Cross-correlation is strongest when selecting one to three additional lags, whereas ATEMPO performs best among the reported selectors at five.

MC\_Maze decoding decodes hand velocity in the first 200\,ms after movement onset from the 20 leading population PCs at the selected lag bins, with no current-bin input, so performance depends entirely on choosing lead times. ATEMPO's phase-conditioned ranking reaches $R^2=0.619$ at five lags where the standard baseline choices collapse. Five evenly spaced lags score $0.626$, showing once redundancy between neighboring bins is avoided, the precise lags matter less on this task, consistent with the interpretation below. 

The official benchmark uses inferred firing rates and a fixed neural lead of approximately 140 ms, whereas our comparison uses smoothed spike counts to isolate the effect of lag selection. These results should therefore be interpreted as comparisons between input-selection strategies, not as state of the art neural decoding results.

\paragraph{Interpretation}
The neural experiments also expose a limitation of cumulative TE differencing. By the chain rule,
\[
\delta_k(l)
=
I\!\left(Y;X_{t-k}\mid
\mathbf{X}^{(k-1)}_t,\mathbf{Y}^{(l)}_t\right).
\]
This measures the information contributed by a lag beyond all shallower source lags, not its standalone predictive value. For strongly autocorrelated inputs, a delayed observation may be individually predictive yet contribute little additional information once nearby observations are conditioned on. Consequently, the largest incremental TE need not identify the best single input feature.

The results show ATEMPO provides a useful information-theoretic ranking of temporal inputs, particularly at larger budgets, but incremental information and downstream predictive utility need not coincide for smooth, redundant signals. This is visible in Table~\ref{tab:nlb_neural} where cross-correlation has an advantage at single-lag budgets and in the strength of evenly spaced lags on MC\_Maze.

\section{Additional Real-World Evaluations}
\label{sec:appendix-realworld}

We further evaluate how ATEMPO can be used to improve downstream forecasting on time series variable selection across five benchmarks.
\subsection{Time series variable selection}
ChronoEpilogi~\citep{chronoepilogi} compares methods for minimal-size subset selection for multivariate time series forecasting, where the resulting minimal-size subset leads to optimal predictive performance of a given target variable.

ATEMPO answers a closely related question with one pairwise fit per candidate: the transfer entropy from the candidate's past 24 lags to the target one step ahead, conditioned on the target's own
24 lags, ranks the candidates by the information they add beyond the target's history. We compare this ranking on all five of ChronoEpologi's benchmark datasets: Electricity, Solar, METR-LA, PEMS-BAY and Traffic. We use their splits, sampling rules and code to compare to ChronoEpilogi (the
forward-equivalent variant) and GroupLasso. We also include two further baselines: ranking by training correlation and choosing random subsets.
\begin{table}[t]
\centering\small
\caption{Comparison to ChronoEpilogi results: test $R^2$ of the ridge forecaster (mean over the 10 targets) with, in parentheses, the number of series each method keeps when choosing its own size on a validation split of the training part; NoSelection uses all series, and the last column is the target's own past alone. $^{\dagger}$Traffic's test period is nonstationary; see the appendix.}
\label{tab:vs_overview}
\begin{tabular}{llllll}
\toprule
 & ATEMPO & ChronoEpilogi & GroupLasso & NoSelection & own past \\
dataset &  &  &  &  &  \\
\midrule
Electricity & \textbf{0.901} (30) & 0.899 (24) & 0.900 (36) & 0.824 (320) & 0.882 \\
METR-LA & \textbf{0.885} (2) & 0.884 (1) & 0.883 (4) & 0.178 (206) & 0.884 \\
Solar & 0.989 (13) & \textbf{0.990} (10) & \textbf{0.990} (10) & 0.988 (136) & 0.988 \\
PEMS-BAY & 0.925 (4) & \textbf{0.927} (5) & \textbf{0.927} (8) & -0.831 (324) & \textbf{0.927} \\
Traffic$^{\dagger}$ & 0.710 (13) & \textbf{0.842} (13) & 0.792 (38) & -1.048 (861) & 0.787 \\
\bottomrule
\end{tabular}
\end{table}
\begin{figure}[t]
  \centering
  \includegraphics[width=\textwidth]{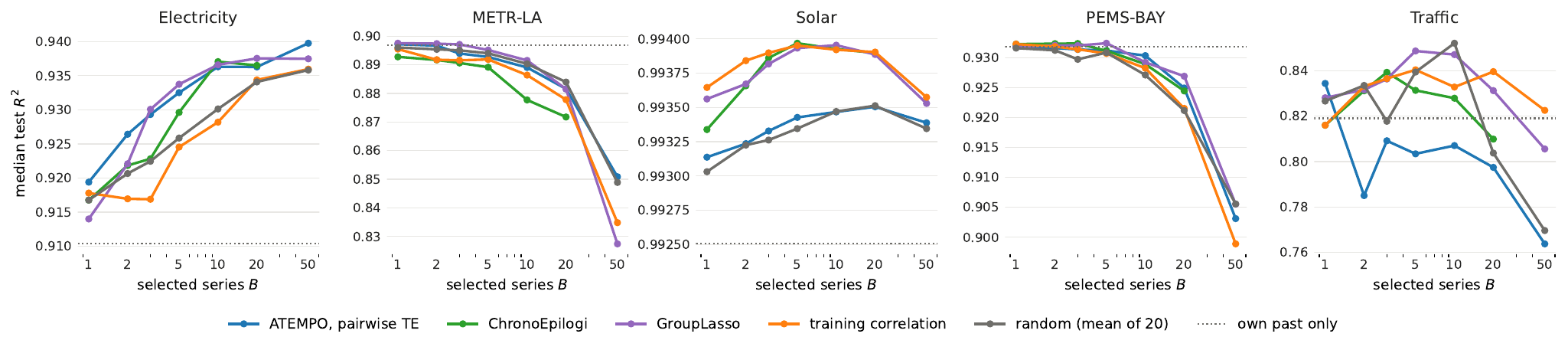}
  \caption{All five datasets: median ridge test $R^2$ against the number of selected series $B$, by selector
  (the greedy wrapper is omitted; dotted line = the target's own past only). Note the vertical scales differ by
  panel: the three saturated datasets span a few thousandths of $R^2$.}
  \label{fig:vs_all}
\end{figure}
We score all baselines using two forecasters: a ridge and a gradient-boosting model on the target's own 24 lags plus 24 lags of each selected series, one step ahead, test $R^2$ on the held-out tail. This differs compared to the more complex models used by~\citep{chronoepilogi} due to computational cost, so we cannot compare \(R^2\) results directly.
The task is defined as in ChronoEpilogi~\citet{chronoepilogi}, each baseline uses the same method to choose its own number of series using the validation split. 
ATEMPO's fits use only the training data split, and three circularly-shifted candidates per target provide a null scale.

Table~\ref{tab:vs_overview} cells report accuracy with the chosen subset's size in parentheses. On METR-LA, Solar and PEMBS-BAY the own-past column shows there is almost nothing beyond the target's history to find, and all three selectors land within $0.003$ of one another while keeping a few percent of the series, so ATEMPO reproduces this headline parsimony result. On Electricity, a ridge given all 320 series ($R^2 = 0.824$, mean) does worse than the target's own past (0.882), so keeping few, good series is better than keeping everything. Figure~\ref{fig:vs_all} shows that for the subset selection task, the ATEMPO ranking is the strongest selector at budget's 1,2 and 50.

On Traffic, ATEMPO's chosen subset degrades performance below the target's own past at smaller budgets that the other baselines. We investigate this result, finding that Traffic's test period (the last 30\%) is strongly nonstationary: with every series standardised on the training part, the panel's weekly standard deviation
sits near 1 throughout training and drifts to 1.2--1.3 over the second half of the test period, variance levels never seen in training. ATEMPO also selects series where this effect is more significant, specifically series 567, present in three of ATEMPO's ten top-5 sets, which has a test standard deviation is
$3.2\times$ its training value. Counting across all ten targets' top-5 sets (50 picks per selector), 4 of ATEMPO's picks have a test-to-train standard-deviation ratio above 2 (series 567 three times, series 512 once), against 1 of 50 for ChronoEpilogi and none for correlation or GroupLasso.

Transfer entropy measures training-period information and carries no protection against candidates whose regime changes after training; the linear selectors' rankings happen to align with what survives the shift. We therefore treat Traffic as a robustness finding about distribution shift rather than a comparison of selection quality, and report
it here for completeness.

\end{document}